\documentclass[preprints,article,accept,moreauthors,pdftex]{Definitions/mdpi} 
\firstpage{1} 
\pubvolume{xx}
\issuenum{1}
\articlenumber{5}
\pubyear{2019}
\copyrightyear{2019}
\history{Received: date; Accepted: date; Published: date}
\pdfoutput=1

\Title{Randomized Projection Learning Method for Dynamic Mode Decomposition}%Approximate Koopman Eigen Spectrum by Randomized Kernelized DMD Methods}
\Author{Sudam Surasinghe $^{1}$ and Erik M.~Bollt $^{2}$}

\AuthorNames{Sudam Surasinghe, Ioannis Kevrekidis and Erik M.~Bollt}

\address{%
$^{1}$ \quad Department of Mathematics, Clarkson University, Potsdam, NY 13699\\
$^{2}$ \quad Electrical and Computer Engineering and $C^3S^2$
the Clarkson Center for Complex Systems Science, Clarkson University, Potsdam, New York 13699}

\abstract{ A data-driven analysis method known as dynamic mode decomposition (DMD) approximates the linear Koopman operator on projected space. In the spirit of Johnson-Lindenstrauss Lemma, we will use random projection to estimate the DMD modes in reduced dimensional space.  In practical applications, snapshots are in high dimensional observable space and the DMD operator matrix is massive. Hence, computing DMD with the full spectrum is infeasible, so our main computational goal is estimating the eigenvalue and eigenvectors of the DMD operator in a projected domain. We will generalize the current algorithm to estimate a projected DMD operator. We focus on a powerful and a simple random projection algorithm that will reduce the computational and storage cost.  While clearly, a random projection simplifies the algorithmic complexity of a detailed optimal projection, as we will show, generally the results can be excellent nonetheless, and quality understood through a well-developed theory of random projections.  We will demonstrate that modes can be calculated for a low cost by the projected data with sufficient dimension. 
}


\begin{document}
%%%%%%%%%%%%%%%%%%%%%%%%%%%%%%%%%%%%%%%%%%

%%%%%%%%%%%%%%%%%%%%%%%%%%%%%%%%%%%%%%%%%%

\section{Introduction}
Modeling real-world phenomena in physical sciences to engineering, videography, economics is limited due to computational costs. Real-world systems require dynamic nonlinear modeling and as Dynamic Mode Decomposition (DMD)\cite{schmid:hal-01020654,RowleyMezic2009,TuDMD} is an emerging tool in this area which can use the data directly rather than intermediary differential equations. However, the data in real-world applications is enormous. The DMD algorithm's reliance on a Singular Value Decomposition(SVD) appears to be a limiting factor due to storing and calculating the SVD of such a matrix. One can notice that SVD calculation of snapshot matrices in big projects would require the use of supercomputers or days of computation. In order to allow processing on small scale computers and in a shorter time frame, we propose to develop an algorithm based on a randomized projection\cite{RP_Gupta_2000} which is used to reduce the dimension of observable space in DMD which we call rDMD.  In order to utilize  and carefully analyzed  the rDMD, we will use the Johnson-Lindenstrauss lemma.  It is clear that a random projection is simple as compared to  a detailed optimal projection method, but our analysis and examples demonstrate nonetheless the quality and efficiency.

Strong theoretical support from Johnson-Lindenstrauss(JL) Lemma \cite{JLlemma} makes the random projection method reliable and has extensive utilization in the field of data science. The JL lemma says that if data points lie in sufficiently high dimensional space, then those data points may be projected into a sufficiently low dimensional space while approximately preserving the distance of the data points.   Furthermore, the projection can be done just by a random matrix which makes algorithms based on JL lemma both past and simple. Hence, this tool is more powerful and adopted heavily in data science. JL lemma based Random projection and SVD based projection can be used to project $N$ dimensional data into lower dimension $L<< N$. Data matrix $X_{N\times M}$ can be projected by random projection into lower dimension ($L$) subspace as $X_L:=RX$ where $R$ is a random matrix with unit length. Hence, the random projection is very simple because it relies only on matrix multiplication. Also computational complexity is $\mathcal{O}(MLN)$ while SVD has computational complexity $\mathcal{O}(NM^2)$ when $M<N$ \cite{RPvSVD}. We will use the random projection to project high dimensional snapshot matrices into  a manageable low dimensional space. In theoretical perspective, the dimension of input and output spaces in the Koopman operator can be reduced by the random projection method thus reducing the storage and computational cost of the DMD algorithm. 

Both DMD and rDMD are grounded in theory through the application of the Koopman operator (Rowley.).  They are numerical methods to estimate the linear  Koopman operator that identifies spatial and temporal patterns from a dynamical system. Theoretical support of the Koopman operator theory makes the these algorithms strong. Our new randomized DMD (rDMD) algorithm targets to address issues that arise in SVD based existing DMD methods by reducing the dimensionality of the matrix just by using matrix multiplication.  rDMD can achieve very accurate results with low-dimensional data embedded in high-dimensional observable space. 
 
 This paper will summarize the Koopman operator theory, existing DMD algorithms, and random projection theory in section (2). Then we will discuss our proposed randomized DMD algorithm in section (3), and finally, in section (4), we will provide examples that support our approach.

\begin{comment}

\begin{center}
\begin{tabular}{ l | c   }
  \hline			
Notation & Meaning\\
  \hline  
  $M$ & Number of data points\\
    $N$ & Number of features\\
    $\mathcal{K}$ & Koopman operator\\
    $\mathcal{F}$ & vector space of observable\\
    $\mathcal{F}_N$ & Finite dimensional approximation of $\mathcal{F}$ \\
    $\hat{\mathcal{K}}$ & Aproximation of Koopman operator\\
     $A$ & any matrix in $\mathbb{R}^{N\times N}$\\
       $\bar{\pmb{A}}$ & Least square solution\\
       $\hat{\bar{\pmb{A}}}$ & Approximation of $\bar{\pmb{A}}$\\
       
\end{tabular}
\end{center}
\end{comment}
\section{Dynamic Mode Decomposition and Background Theory }
The focus of this paper is to approximate the eigenpairs of the Koopman operator based on the randomized dynamic mode decomposition method. We will first review the underlining theory about the Koopman operator. 
\subsection{Koopman Operator}
\begin{figure}
      \centering
       \includegraphics[width=0.9\textwidth]{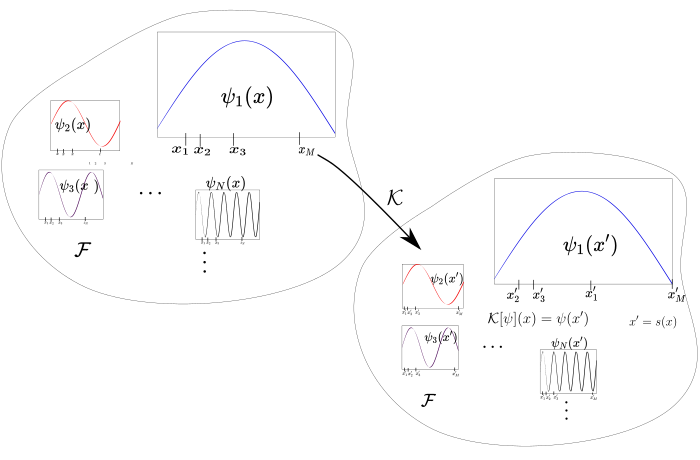}\caption{This figure shows the behavior of Koopman operator $\mathcal{K}$ in observable space $\mathcal{F}$ associated with a dynamical system $S$. The Koopman operator evaluate the observable $\psi$ at downstream or future $x'=x_{n+1}=s(x_n)$. }
        \label{fig:Koopman}
    \end{figure}

Consider a discrete-time dynamical system
\begin{align}\label{eq:dynamic}
    x_{n+1}=S(x_n),
\end{align}

where $S:\mathcal{M}\to \mathcal{M}$ and $\mathcal{M}$ is a finite dimensional manifold. (If we have a differential equation or continuous time dynamical system, flow map can be considered.) The variable $x$ is often recognized as a state variable and $\mathcal{M}$ as phase space. The associated Koopman operator is described as the evaluation of observable functions (Fig.~\ref{fig:Koopman}) $\psi: \mathcal{M} \to \mathbb{R}$
in function space $\mathcal{F}$. Instead of analysing the individual trajectories in phase space, the Koopman operator operates on the observations \cite{koopman1931,BOLLTGeoDMD,TuDMD,RowleyMezic2009}. 
\begin{Definition}[Koopman operator \cite{koopman1931}]\label{Def:Koop}
The Koopman operator $\mathcal{K}$ for a map $S$ is defined as  following composition,
\begin{align}
    \mathcal{K} &:\mathcal{F}\rightarrow \mathcal{F}\nonumber \\
    \psi \mapsto \mathcal{K}[\psi]  &=\psi\circ S
\end{align}
on the function space $\mathcal{F}$.
\end{Definition}
It is straight forward to prove \cite{koopman1931}, 
\begin{align}
    \mathcal{K}[a\psi_1+b\psi_2]=a (\psi_1\circ S) + b (\psi_2\circ S) =a \mathcal{K}[\psi_1]+b\mathcal{K}[\psi_2]
\end{align}
for $\psi_1,\psi_2\in\mathcal{F}$ and $a,b\in \mathbb{C}$, and therefore the Koopman operator is linear on $\mathcal{F}$. This is interesting and important property of the operator because the associated map $S$ most probably will be non-linear.  Even though the operator is associated with a map that evolves in finite dimensional space, $\mathcal{F}$ the function space which the operator acts on could possibly be an infinite dimensional. This is the trade-off between cost for the linearity \cite{BOLLTGeoDMD}. %For practical consideration, we need to approximated the operator on finite dimensional projected observable space $\mathcal{F}_N$  (See details in \cite{Korda_2017}). We can define $\mathcal{F}_N:=\Span \{\psi_1,\ \psi_2,\dots,\ \psi_N\}$ for given set of linearly independent basis functions $\psi_i\in\mathcal{F},\ i=1,\ \dots,\ N$ and then Koopman operator $\mathcal{K}$ can be approximated by $\hat{\mathcal{K}}:\mathcal{F}_N \rightarrow \mathcal{F}_N$ with sampling at $M$ points in the state space. 

Spectral analysis of the Koopman operator can be used to decompose the dynamics which becomes the key success in the DMD. Assuming the spectrum  of Koopman operator $\mathcal{K}$ is given by
\begin{align}\label{eq:specKoop}
    \mathcal{K}\psi_i(x)=\lambda_i\psi_i(x)~~~ i=1,\ 2,\ 3,\ \dots
\end{align}
then vector-valued observables $\pmb{g}:\mathcal{M}\to \mathbb{R}^N$(or $\mathbb{C}^N$) can be represented by
\begin{align}\label{eq:kopObsDec}
    \pmb{g(x)}=\sum_{i=1}^{\infty}\psi_i(\pmb{x})\pmb{\phi_i},
\end{align} where $\pmb{\phi_i}\in\mathbb{R}^N$(or $\mathbb{C}^N$) are the vector coefficients of the expansion and called ``Koopman modes''(here we assumed that components of $\pmb{g}$ lie within the span of the eigenfunctions
of $\mathcal{K}$).  Note that the observable value at time $n+1$ is given by 
\begin{align}
    \pmb{g(x_{n+1})}=\sum_{i=1}^{\infty}\lambda_i^n \psi_i(\pmb{x_0})\pmb{\phi_i}.
\end{align} 
This decomposition can be used to separate the spacial and time components of the dynamical system and can be used to isolate the specific dynamics.

\subsection{Dynamic Mode Decomposition}\label{DMD}
The dynamic mode decomposition is a data-driven method to estimate the Koopman modes from numerical or experimental data \cite{RowleyMezic2009}. % and few slight variant of DMD algorithms are available defending on the numerical process of estimating those Koopman eigenpairs (Ref. Schmid, Tu.).  We will summarize more numerically stable exact DMD algorithm (Ref. Tu) in rest of this section.  
Suppose dynamics are govern by eq.~(\ref{eq:dynamic}) for any state $\pmb{x}$ and vector valued measurements are given by observable $\pmb{g(x)}\in \mathbb{R}^N$. For a given set of data $X=[\pmb{g(x_0)}\ \pmb{g(x_1)}\ \dots\ \pmb{g(x_{M-1})} ],\ Y=[\pmb{y_0}\ \pmb{y_1}\ \dots\ \pmb{y_{M-1}} ]$ where $\pmb{y_i}=\pmb{g(s(x_i))}$, the Koopman modes and eigenvalues of the Koopman operator can be estimated through solving the least-squares problem
\begin{align}\label{lseKoop}
\mathbb{K}=\argmin_{K} ||KX-Y||_F^2=\argmin_{K}\sum_{i=0}^{M-1}||K\pmb{g(x_i)}-\pmb{y_i}||_2^2
\end{align}
    
%The operator $\hat{\mathcal{K}}$ (EDMD operator) is defined by 
%\begin{align}\label{eDmdKoop}
%\hat{\mathcal{K}} g = %c^{H}\bar{\pmb{A}}\psi
%\end{align}
%for any $g=c^H\psi,\ c\in \mathbb{C}^N$, $c^H$ means the conjugate transpose of the matrix $c$, and $\bar{\pmb{A}}=\Psi(Y)\Psi(X)^{\dagger}$.
and $\mathbb{K}=YX^{\dagger}$ (here $X^{\dagger}$ is the pseudo-inverse of $X$) is defined as the "Exact DMD" operator \cite{TuDMD}. The eigenvalue ($\hat{\lambda}$) of $\mathbb{K}$   is an approximation of an eigenvalue ($\lambda$) of $\mathcal{K}$; the corresponding right  eigenvector($\hat{\phi}$) is called the DMD mode and approximates the Koopman mode ($\phi$). Then the observable value $\pmb{g(x(t))}$ at time $t$ can be modeled as 
\begin{align}\label{eq:OrgDec}
    \pmb{g(x(t))}=\sum_{i=1}^{r}\psi_i(\pmb{x_0})\hat{\pmb{\phi}}_i \hat{\lambda}_i^t
\end{align}
where $r$ is the number of selected DMD modes and demonstrates the finite dimensional approximation for vector-valued observable $\pmb{g}$ under the Koopman operator. Based on this decomposition, data matrices can be expressed as
\begin{align}\label{eq:dataDecomDMD}
    X_{N\times M}&= \Phi_{N\times r} T_{r \times M} \\ \nonumber
    Y_{N \times M}&= \Phi_{N\times r}\Lambda_{r \times r} T_{r \times M}
\end{align}
where $\Phi=[\psi_1(x_0)\pmb{\phi}_1~~\psi_2(x_0)\pmb{\phi}_2~ \dots ~\psi_r(x_0)\pmb{\phi}_r]$, $T$ is a Vandermonde matrix with $T_{ij}=\lambda_i^{j-1}$ for $i=1,2,\dots r$, $j=1,2,\dots, M$  and $\Lambda=diag\{\lambda_1,~\lambda_2,\dots, \lambda_r\}$. Note that with the above decomposition $\mathbb{K}=YX^{\dagger}=\Phi\Lambda T T^{\dagger}\Phi^{\dagger}$. We will suppose $\mathbb{K}$ has distinct eigenvalues $\lambda_i$, columns of $X$ are linearly independent and $r\le M$.   
 In practical applications, we are expected to fully understand the data set by relatively few ($r<<M$) modes. This can be considered as one of the dimension reduction steps of the algorithm. Additionally, dimension of columns of the data matrix need to be reduced.

In practice, the columns of data matrix $X$ (and $Y$) are constructed by the snapshot matrices of spatial observable data. More often, those snapshots lie in high dimension space $\mathbb{R}^N$ ($N>>1$ and roughly $\mathcal{O}(10^{15})$ to $\mathcal{O}(10^{10})$), but the number of snapshots or time steps ($M$) are small and often it is $\mathcal{O}(10^{3})$ to $\mathcal{O}(10^{1})$ \cite{chenTuRowley2012}. Hence, computing the spectrum of matrix $\mathbb{K}$ is infeasible even though most of the eigenvalues will be zero. Therefore we can project our data matrices $X, Y$ into a low dimensional space $R^{L}$ with $r\le L\le M<< N$, therefore need to estimate the spectrum of $\mathbb{K}$ based on the computation on projected space. Our proposed rDMD method is focused on this dimension reduction step.   

In the next section (Sec.~(\ref{sec:rDMD_main})) we will discuss more details about the calculation. Note that current methods are based on the singular value decomposition of the data matrix $X$ to construct a projection and our proposed algorithm is based on the random projection method to project data into a low dimensional space.

\subsection{Random projection}
The random projection method is based on the Johnson-Lindenstrauss lemma which is applied by many data analysis methods. 

\begin{Theorem}[Johnson-Lindenstrauss Lemma\cite{JLlemma}] For any $0<\epsilon<1$ and any integer $M>1,$ let $L$ be a positive integer such that $L\ge L_0$ with $L_0=\frac{C\ln M}{\epsilon^2},$ where $C$ is a suitable constant ($C\approx 8$ in practice,$C=2$ is good enough). Then for any set $X$ of $M$ data points in $\mathbb{R}^{N}$, there exists a map $f: \mathbb{R}^N \to \mathbb{R}^L$ such that
for all $x_1,x_2 \in X$, \[(1-\epsilon)||x
_1-x_2||^2\le ||f(x
_1)-f(x_2)||^2\le (1+\epsilon)||x
_1-x_2||^2. \]
\end{Theorem}
\begin{Theorem}[Random Projection \cite{RP_Gupta_2000}] For any $0<\epsilon,\delta <\frac{1}{2}$ and positive integer $N$, there exists a random matrix of $B$ of size $L\times N$ such that for $L\ge L_0$ with $L_0=\frac{C \ln(1/\delta)}{\epsilon^2}$. and
for any unit-length vector $x\in R^N$
\[Pr\{|||Bx||^2-1|>\epsilon\}\le \delta\]
or
\[Pr\{|||Bx||^2-1|>\epsilon\}\le e^{-CL\epsilon^2}\]
\end{Theorem}

A low rank approximation for both $X, Y$ can be found using the random projection method. Notice that both these matrices have $M$ points from $N$ dimensional observable space and therefore we can use random projection matrix $B$ of size $L\times N$ with the $L\ge \frac{C\ln N}{\epsilon^2}$ which provides $\epsilon-$ isometry to $\mathbb{R}^N$. (See Fig.~\ref{fig:rkDMDbigPic} for details).

\section{Randomized Dynamic Mode Decomposition}\label{sec:rDMD_main}  
In this section we will generalize currently used DMD algorithms and then we will discuss our proposed randomized DMD algorithm.

\subsection{DMD on projected space}\label{DMD_on_Pro}
\begin{figure}[htb]
        \centering
        \includegraphics[width=0.6\textwidth]{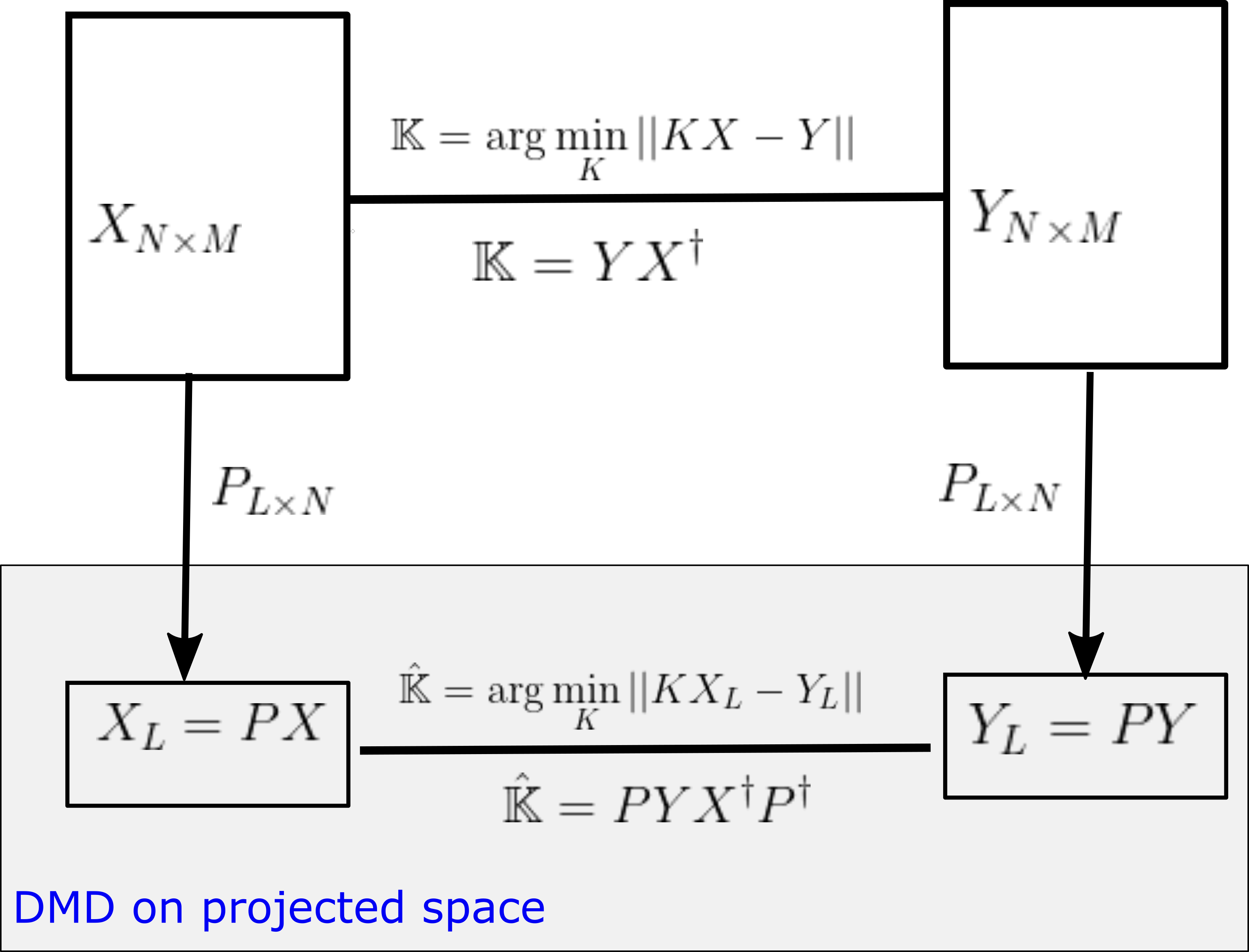}\caption{DMD operator on projected space. This figure shows the relationship between DMD operator on original space and DMD on projected space. The operator $\mathbb{\hat{K}}$ on projected space is defined in Eq.~(\ref{proDMD1}) and can be calculated by Eq.~(\ref{proDMD1}).}
        \label{fig:rkDMDbigPic}
    \end{figure}

As we mentioned in section \ref{DMD}, computational and storage cost of DMD can be reduced by projecting data into a low dimensional observable space. Let $P\in \mathbb{R}^{L\times N}$ be any rank $L$ projection matrix, then dimension of data matrices $X,\ Y\in \mathbb{R}^{N\times M}$ can be reduced to $L \times M$ by the projection $X_L=PX,\ Y_L=PY$. The DMD operator on projected space (see Fig.~(\ref{fig:rkDMDbigPic})) is given by,
\begin{align}\label{proDMD1}
    \hat{\mathbb{K}}=\argmin_{K\in \mathbb{R}^{L\times L}} ||KX_L-Y_L||_F^2 =\argmin_{K\in \mathbb{R}^{L\times L}} ||KPX-PY||_F^2
\end{align}
and $\hat{\mathbb{K}}=PY(PX)^{\dagger}$. Therefor
\begin{align}\label{proDMD}
    \hat{\mathbb{K}}=PY(PX)^{\dagger}=PYX^{\dagger}P^{\dagger}=P\mathbb{K}P^{\dagger}
\end{align}
where $\mathbb{K}=YX^{\dagger}$ is the DMD operator on original space.

\begin{Proposition}
Some eigenpairs $(\lambda, \phi)$ of $\mathbb{K}$ can be obtain by $(\lambda_L, \phi_L)$ of projected DMD $\hat{\mathbb{K}}$ with $\lambda=\lambda_L$ and $\phi=P^{\dagger}\phi_L$.
\end{Proposition}
\begin{proof}
Let $(\lambda_L, \phi_L)$ be an eigenpair of $\hat{\mathbb{K}}$. Then $\hat{\mathbb{K}}\phi_L=\lambda_L\phi_L$ and by eq.(\ref{proDMD}), $P\mathbb{K}P^{\dagger}\phi_L=\lambda_L\phi_L$. Now let $P^{\dagger}\phi_L=\phi$, then $\phi_L=P\phi$ because $PP^{\dagger}=I$. Hence $P\mathbb{K}P^{\dagger}\phi_L=P\mathbb{K}\phi=\lambda_LP\phi$ and  $P(\mathbb{K}\phi-\lambda_L\phi)=0$. Since  $\mathbb{K}\phi-\lambda_L\phi=0$ is a solution to the above equation,  $\lambda_L$ is an eigenvalue and the corresponding eigenvector is $\phi=P^{\dagger}\phi_L$ of $\mathbb{K}$.
\end{proof}
 In other words, we can lift up the dimension of eigenvectors in projected space by $P^{\dagger}$ to obtain an eigenvector in original data space. However to avoid the direct calculation of the pseudo-inverse of projection matrix , we can calculate the eigenvector in output space $Y_L$ of the DMD operator and lift up the vector into the original output space $Y$. We can easily show that $\hat{\phi}= Y(PX)^{\dagger}\phi_L$ is an eigenvector of $\mathbb{K}$ for corresponding non-zero eigenvalues.
\begin{align}
    \mathbb{K}\hat{\phi}&= \mathbb{K}Y(PX)^{\dagger}\phi_L=\mathbb{K}YX^{\dagger}P^{\dagger}\phi_L  \\\nonumber
    &=\mathbb{K}\mathbb{K}P^{\dagger}\phi_L \\\nonumber
    &=\mathbb{K}\lambda P^{\dagger}\phi_L \\\nonumber
    &=\lambda YX^{\dagger} P^{\dagger}\phi_L \\\nonumber
    &=\lambda Y(PX)^{\dagger}\phi_L \\\nonumber
    &=\lambda \hat{\phi} \\\nonumber
\end{align}

\begin{figure}[htb]
        \centering
        \includegraphics[width=0.5\textwidth]{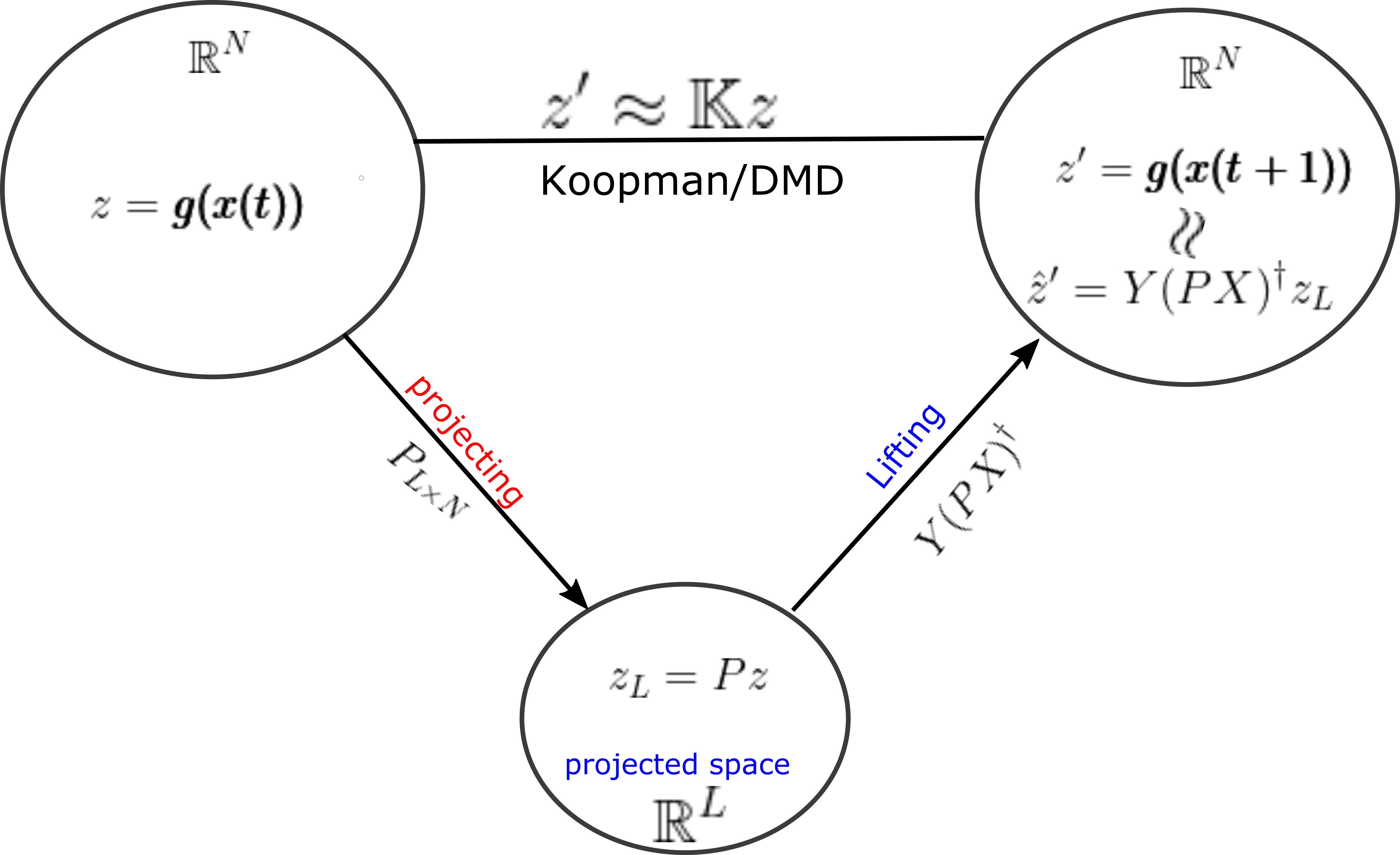}\caption{ The figure shows the projecting operator $P$ and DMD related lifting operator $Y(PX)^{\dagger}=\mathbb{K}P^{\dagger}$ which should be used in DMD algorithms. Instead of using $P^{\dagger}$ as the lifting operator, $Y(PX)^{\dagger}$ can be used for efficient calculations. Moreover, notice that $\hat{z}'_L:=P\hat{z}'=\hat{\mathbb{K}}z_L$.  }
        \label{fig:proAlgo}
    \end{figure}

 Also notice, $P\hat{\phi}=PY(PX)^{\dagger}\phi_L=\hat{\mathbb{K}}\phi_L$ and therefor $\hat{\phi}$ estimate the eigenvector on output space $Y$. Detailed view of this lifting operator is shown by Fig.~(\ref{fig:proAlgo}). It provides the relationship of the lifting operator with the DMD operator acted on any general observable vector $z=\pmb{g(x(t))}\in \mathbb{R}^N$. 
 
 Next, the focus move to the spatial-temporal decomposition of projected data matrices by spectrum of the DMD operator. Note that observable value $\pmb{g(x(t))}$ at time $t$ can be modeled as $\pmb{g(x(t))}=\sum_{i=1}^{r}\psi_i(\pmb{x_0})P^{\dagger}\pmb{(\phi_L)}_i \hat{\lambda}_i^{t}$ and 
 similar to the eq.~(\ref{eq:dataDecomDMD}), data can be decomposed as
 \begin{align}\label{eq:ProDmdVarSep}
     X_{N\times M} &= P_{N\times L}^{\dagger}\Tilde{\Phi}_{L\times r} T_{r\times M}\\ \nonumber
   Y_{N\times M}&= P_{N\times L}^{\dagger}\Tilde{\Phi}_{L\times r} \Lambda_{r\times r} T_{r\times M}.
 \end{align}
 This decomposition leads to $\mathbb{K}=YX^{\dagger}=P^{\dagger}\Tilde{\Phi}\Lambda T T^{\dagger}\Tilde{\Phi}^{\dagger}P$ and if $r\le L$ all the non-zero eigenvalues and corresponding eigenvectors of $\mathbb{K}$ can be constructed by the projected DMD operator. Further, eq.~(\ref{eq:ProDmdVarSep}) can be use to isolate the spatial profile of interesting dynamical compotes such as attractors, periodic behaviors, etc.

 Based on the choice of the projection matrix we will have alternative ways to estimated the spectrum of the DMD operator. 
\begin{remark}[Projection by SVD]
Commonly used projection matrix is based on SVD of the input matrix $X=U\Sigma V^*$ and projection matrix is chosen to be $P=U^*$, here $^*$ represents the conjugate transpose of a matrix. Using eq.~\ref{proDMD} and SVD of X, the operator on projected space can be formulated as  $\hat{\mathbb{K}}=U^{*}YV\Sigma^{-1}$. 
\end{remark}
\begin{remark}[Standard DMD and Exact DMD]
 Let eigenpair of a SVD based $\hat{\mathbb{K}}=U^{*}YV\Sigma^{-1}$ be given by $(\lambda,\phi_L)$. In standard DMD (Ref. Schmid Paper\& Tu paper) use the eigenvector $P^{\dagger}\phi_L=U\phi_L$ to estimate eigenvectors of $\mathbb{K}$. On the other hand, in exact DMD(ref. Tu paper) this eigenvector is estimated by $Y(PX)^{\dagger}\phi_L=YV\Sigma^{-1}\phi_L$.
\end{remark}
\begin{remark}
QR decomposition based projection methods on both input and output data $[X\ Y]$ can be used \cite{TuDMD}. 
\end{remark}
 In this paper we are proposing a simple random projection based method to estimate the spectrum of the DMD operator. 
\subsection{Randomized Methods}
Our suggested randomized Dynamic Mode Decomposition(rDMD) is based on the random projection applied to the theory of DMD on projected space. We can reduce the dimension of data matrix $X,\ Y$ in DMD by using a random projection matrix $R_{L\times N }$. In other words we will construct a projection matrix $P$ discussed in sec.~\ref{DMD_on_Pro} as a random matrix $R$ whose columns have unit length and entries  are selected independently and identically from a probability distribution. Therefore, the rDMD matrix on the projected space is given by $\hat{\mathbb{K}}=RY(RX)^{\dagger}$, and if an eigenpair of $\hat{\mathbb{K}}$ is given by $(\lambda, \phi_L)$, then the eigenpair of $\mathbb{K}$ is given $(\lambda, Y(RX)^{\dagger}\phi_L)$. Algorithm \ref{Al:Algo1} represents the major steps needed to estimate the eigenvalues and corresponding eigenvectors of the DMD operator with random the projection method. 

The calculation of the projection matrix of a standard or exact DMD algorithm based on SVD of snapshot matrix $X$  is needed to store a full high resolution data matrix which leads to memory issues. Our proposed rDMD algorithm can avoid these storage issues, because low dimensional matrices $X_L,\ Y_L$ obtained by  matrix multiplications only need to store one row and one column of each matrix at a time. Additionally, this algorithm reduces the computational cost since we only need to calculate pseudo-inverse of comparatively lower dimensional matrix. Choice of the distribution of $R$ can further reduce the computational cost \cite{ACHLIOPTAS2003671}. 
 
\begin{algorithm}[htb]
\SetAlgoLined
\KwData{$X,\ Y\in \mathbb{R}^{N\times M}$}
\KwInput{$\epsilon$}
$L_0=\frac{C\ln M}{\epsilon^2}$\;
Choose $L$ such that %$L=\binom{q+d}{d}\ge L_0$\;
$L\ge L_0$\;
Construct a random matrix $R=\frac{1}{\sqrt{L}}(r_{ij})\in \mathbb{R}^{L\times N}$ such that $r_{ij}~N(0,1)$\;
%Use $k(x,y)=(1+x^Ty)^q$ to find the matrices $G, Q$ such that $G_{ij}=k(b_i,x_j)$ and $Q_{ij}=k(b_i,y_j)$ \;
Calculate $X_L:=RX,\ Y_L:=RY$ \;
Calculate $\hat{\mathbb{K}}=Y_LX_L^{\dagger}$\;
[$\lambda$ $\ \Phi_L$]=eigs($\hat{\mathbb{K}}$)\;
 \KwResult{$diag(\Lambda)$, $YX_L^{\dagger}\Phi_L$}
 \caption{Randomized DMD(rDMD)}\label{Al:Algo1}
\end{algorithm}

One time step forecasting error for any given snapshot by using rDMD algorithm can be bounded by using the JL theory.

\begin{Proposition}[Error Bound]
Let $z=\pmb{g(x(t))},\ z'=\pmb{g(x(t+1))}\in \mathbb{R}^{N}$. Error bound of estimating $z'$ by using the rDMD as $\hat{z}'=YX^{\dagger}R^{\dagger}Rz$ is given by 
\begin{align}\label{eq:rDMDErr}
   E(z';L):=||z'-\hat{z}'||\le \frac{||Rz'-\hat{\mathbb{K}}Rz||}{1-\epsilon}:= UB
\end{align}
with at least the probability of $\mathcal{O}(1/M^2)$ for any $0<\epsilon<1$ with $L>\frac{C\log(M)}{\epsilon^2}$.
\end{Proposition}

\begin{proof}
Since $\hat{\mathbb{K}}=RYX^{\dagger}R^{\dagger}$, the rDMD acts on the projected vector which can be rearranged as $\hat{\mathbb{K}}Rz=R\hat{z}'$. Therefore, 
\begin{align*}
    ||Rz'-\hat{\mathbb{K}}Rz||=||Rz'-R\hat{z}'||.
\end{align*}
Now we can apply the JL theory to attain the  desired error bound.
\begin{align*}
(1-\epsilon)||z'-\hat{z}'|| \le ||Rz'-R\hat{z}'||=||Rz'-\hat{\mathbb{K}}Rz||.
\end{align*}
Hence $||z'-\hat{z}'||\le \frac{||Rz'-\hat{\mathbb{K}}Rz||}{1-\epsilon}$.
\end{proof}

%So far we focused on the direct forecast and now we will consider the final model with decomposition of the data.    

\section{Results and Discussion}
In this section we will demonstrate the theory of rDMD  with a few examples. The first two examples consider the computation for known dynamics and demonstrate the error analysis. The final example will demonstrate application in the field of oceanography and isolate the interesting features by rDMD and compere with the resulting modes with the exact DMD results. 
\subsection{Logistic Map}
\begin{figure}[htb]
        \centering
     \begin{subfigure}[b]{0.45\textwidth}
      \centering
        \includegraphics[width=\textwidth]{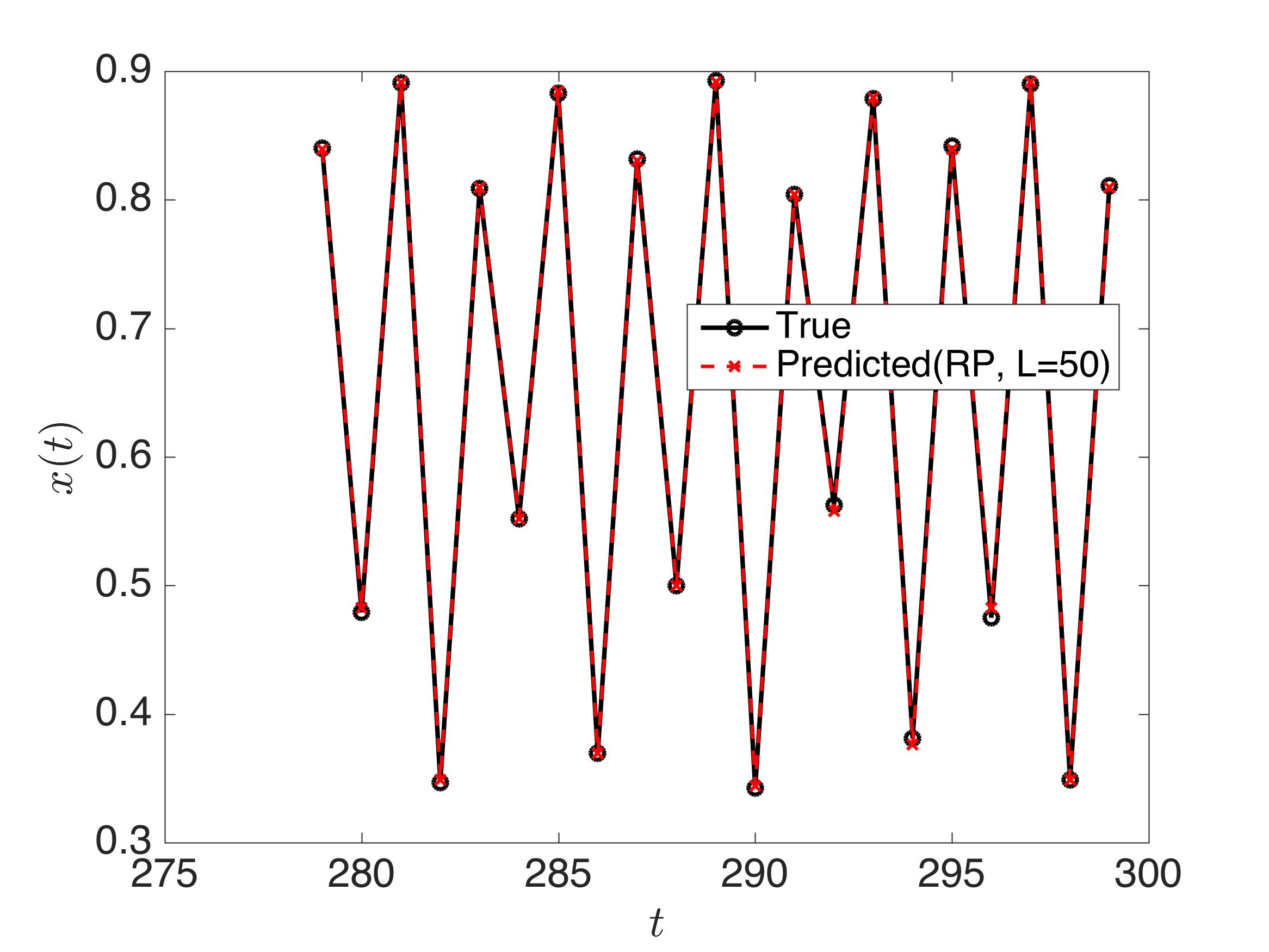}
     \end{subfigure}
          \begin{subfigure}[b]{0.45\textwidth}
      \centering
        \includegraphics[width=\textwidth]{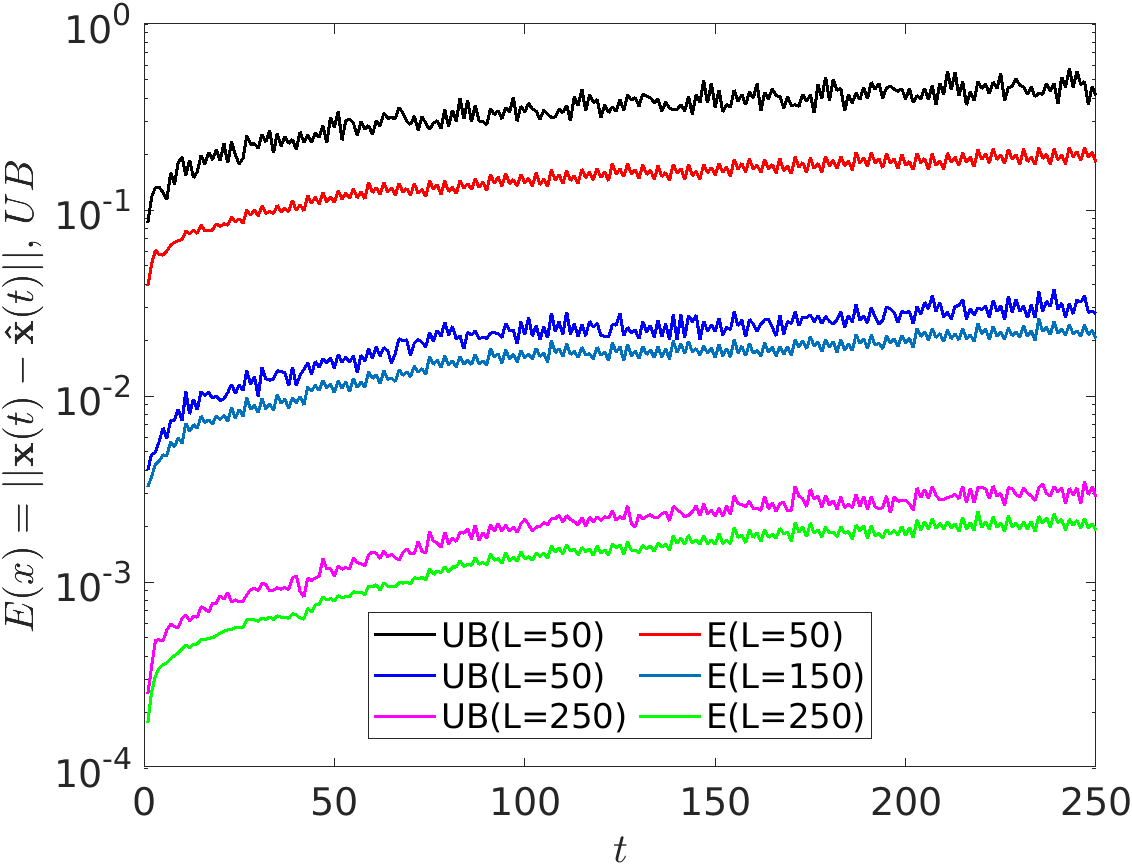}
     \end{subfigure}
     \caption{ (Left) shows the predicted data using the rDMD algorithm with projected dimension $L=50$ compared to the original data from logistic map $x_{n+1}=3.56994x_n(1-x_n)$ for initial condition $x(0)=0.4967$. Further, (Right) figure shows the estimated error and theoretical upper bounds (Eq.~\ref{eq:rDMDErr}) for the some projected dimension $L$ and this example validates the theoretical bound.  }\label{log:OrVsrDMD}
    \end{figure}

\begin{figure}[htb]
        \centering
     \begin{subfigure}[b]{0.45\textwidth}
      \centering
        \includegraphics[width=\textwidth]{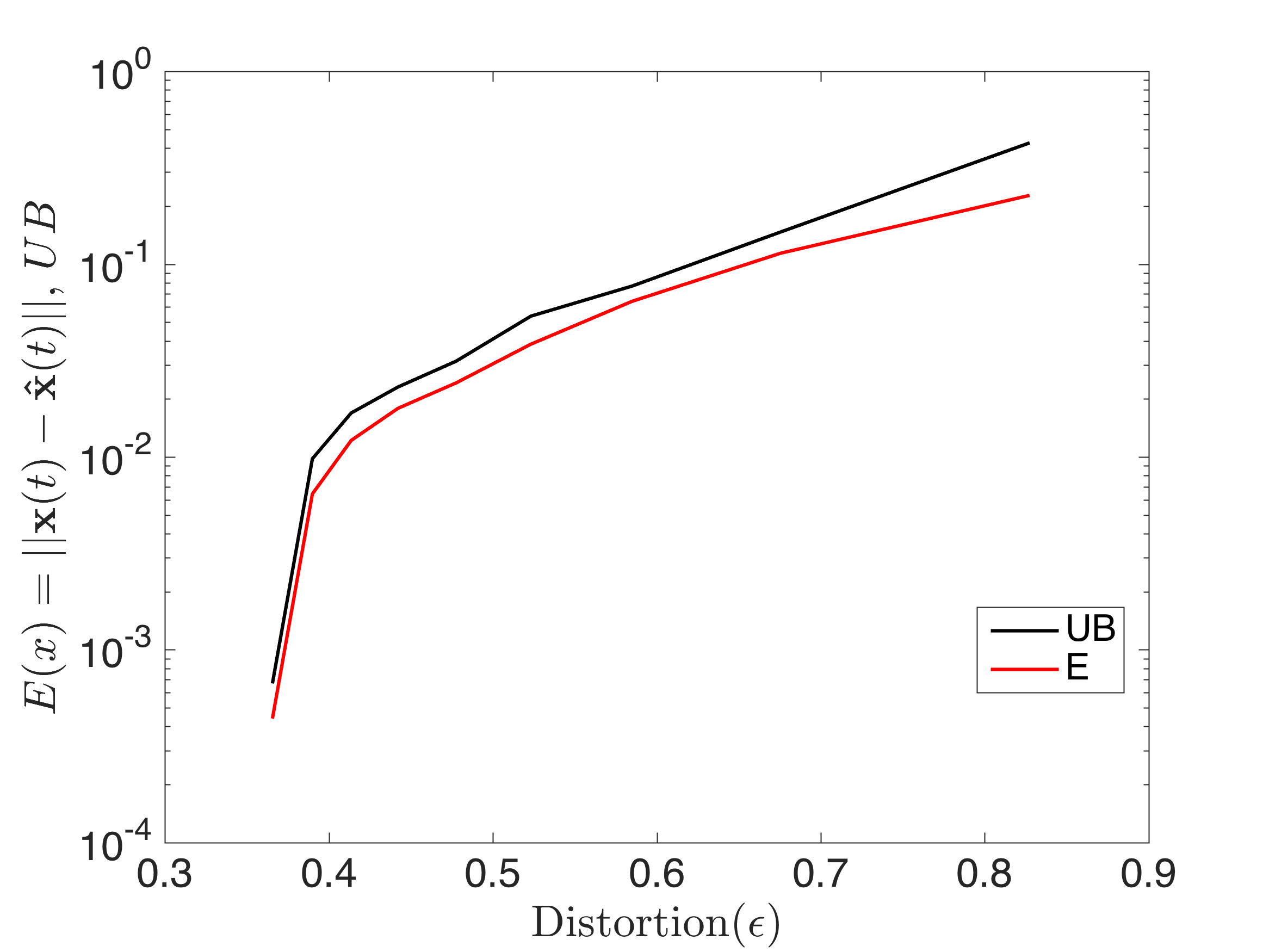}\caption{ }
     \end{subfigure}
          \begin{subfigure}[b]{0.45\textwidth}
      \centering
        \includegraphics[width=\textwidth]{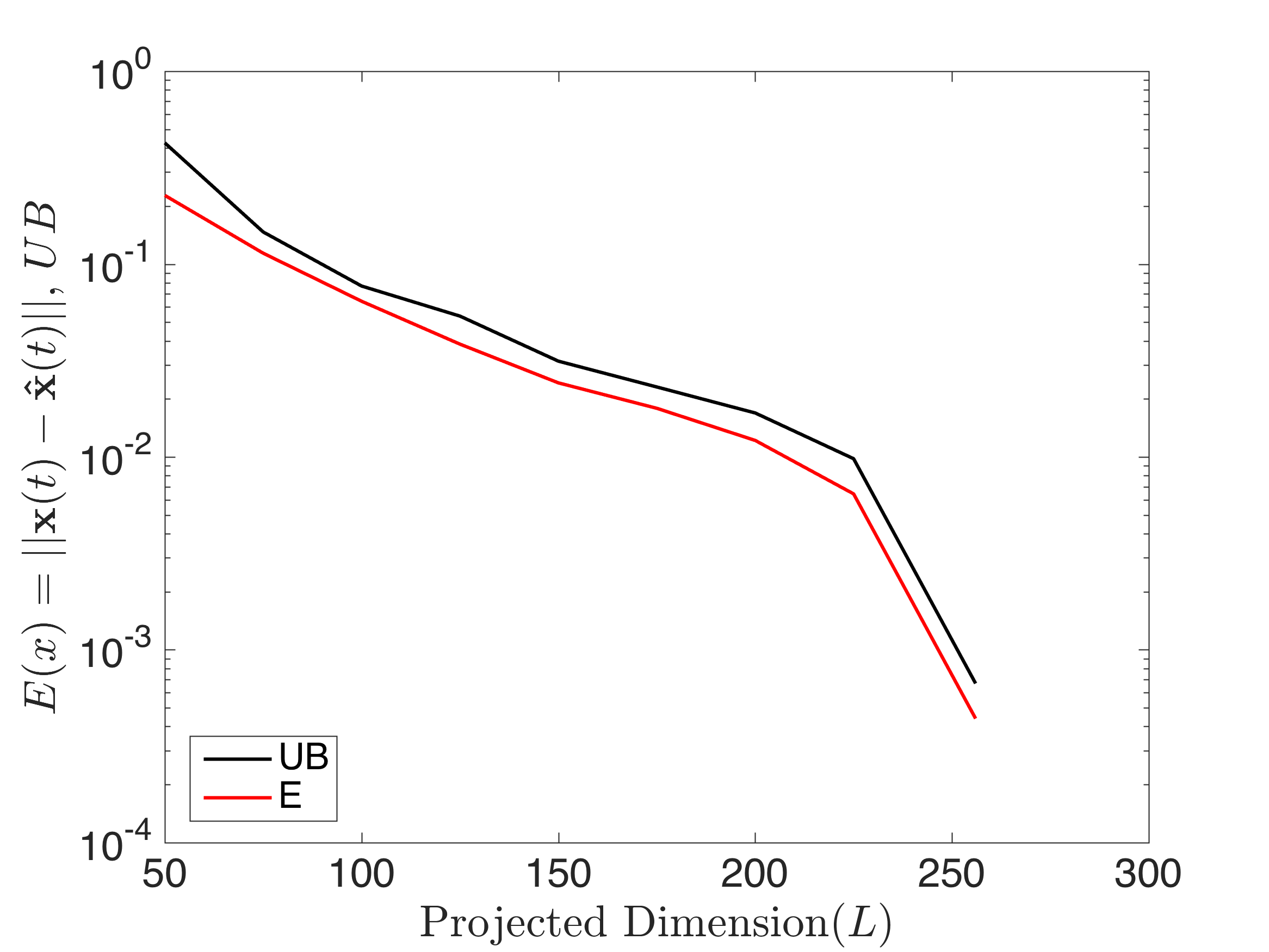}\caption{ }
     \end{subfigure}
     \caption{ (a),(b) shows the prediction error of logistic map $x_{n+1}=3.56994x_n(1-x_n)$ by rDMD and its theoretical upper bounds(Eq.~\ref{eq:rDMDErr}). Figure (a) represents the error with respect to the distortion $\epsilon$ and (b) shows the error with dimenstion of the projected space that will guarantee the bound for this example.}\label{log:Error}
    \end{figure}

We will first consider a dataset of 300 snapshots from a logistic map,
\begin{align*}
    x_{n+1}=ax_n(1-x_n)
\end{align*}
with $a=3.56994$. In this case all the initial conditions will converge to a period-256 orbit. Therefore rank of the snapshot matrix with relatively high samples should be 256.   We forecast the data by using the rDMD method and then analyzed the error of the prediction and compared it with the theoretical upper bound. With $N=5000$ initial conditions and $M=300$ samples, the dimension $L$ of the projecting space can be chosen as $L\ge \frac{C \ln (300)}{\epsilon^2}\approx \frac{34.22}{\epsilon^2}$ when $C=6$. rDMD with projection into 50 dimensional space can accurately forecast the time series data. (Fig.~(\ref{log:OrVsrDMD}) shows the original vs predicted data for one trajectory.) Furthermore, fig.~(\ref{log:Error}) demonstrates the bound of the error of forecast explained in eq.~(\ref{eq:rDMDErr}) and how error relates with the distortion parameter $\epsilon$ (fig.~(\ref{log:Error}) (a)) and dimension of the projected space ((fig.~(\ref{log:Error}) (b))). Since the rank of the snapshot matrix is $256$, any $L\ge 256$ will perform very accurately. This example validates the error bound we discussed in Eq.~(\ref{eq:rDMDErr}) and error of the prediction depends on the error exhibited by the projected DMD operator and the distortion parameter ($\epsilon$ or the projected dimension) from the JL theory.
    
    \subsection{Toy Example: Demonstrates the Variable Separation and isolating dynamics.}\label{sec:toyEx}
    To demonstrate the variable separation and to isolate the spatial structures based on the time dynamics, we consider a toy example(motivated by the example in \cite{DMDbook}) ,
    \begin{align}\label{eq:sechPro}
        z(x,t)=\sum_{j=1}^{20}j\sech(0.1x+j)e^{i\gamma_jt}=\sum_{j=1}^{20}\Phi_j(x)T_j(t)
    \end{align}
    where $\gamma_j$'s are constants, and let $\Phi_j(x)=j\sech(0.1x+j)$ and $T_j(t)=e^{i\gamma_jt}$.
    
            \begin{figure}[htb]
        \centering
         \begin{subfigure}[b]{0.32\textwidth}
      \centering
        \includegraphics[width=\textwidth]{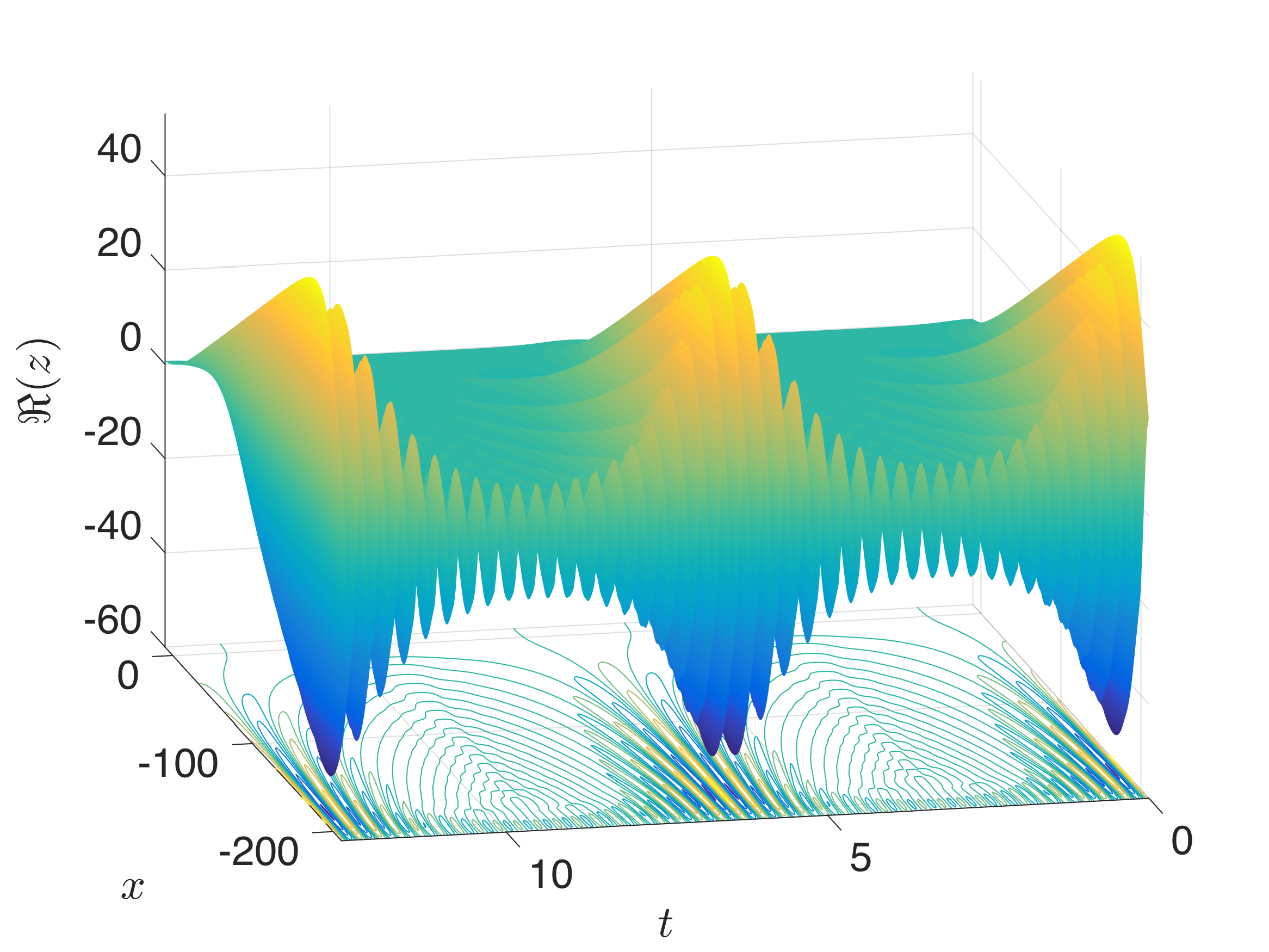}
        \end{subfigure}
                 \begin{subfigure}[b]{0.32\textwidth}
      \centering
        \includegraphics[width=\textwidth]{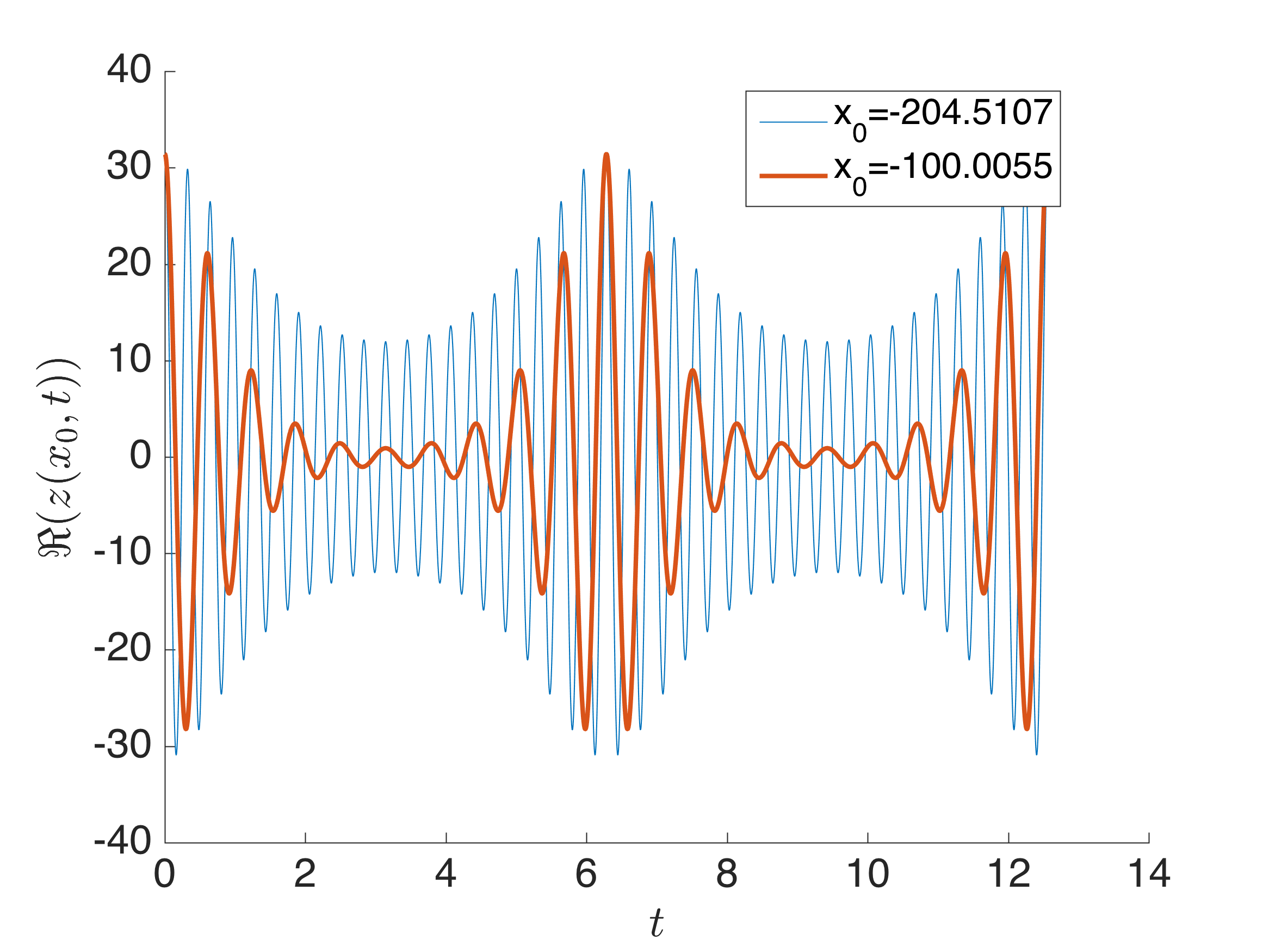}
        \end{subfigure}
        \begin{subfigure}[b]{0.32\textwidth}
      \centering
        \includegraphics[width=\textwidth]{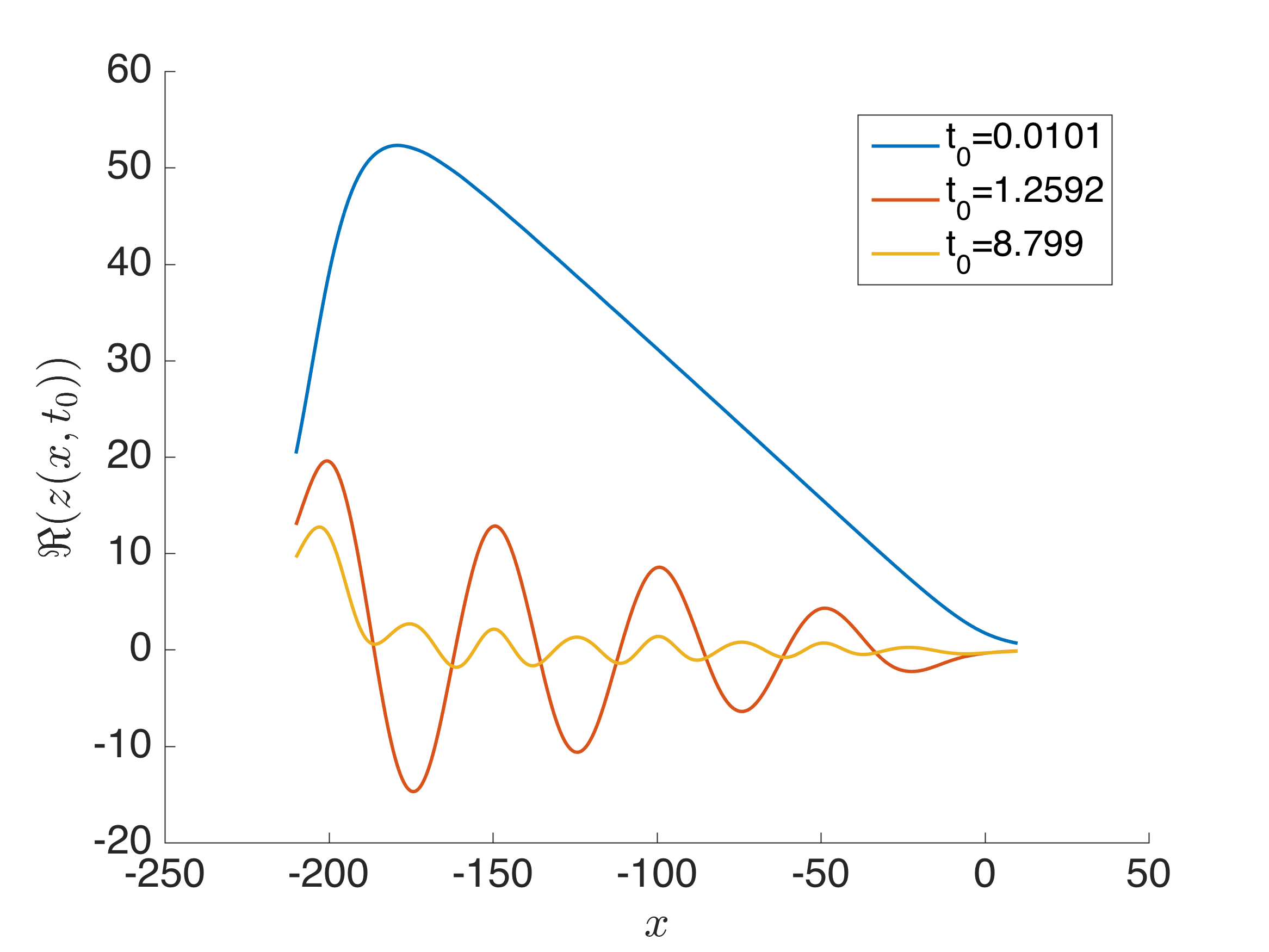}
        \end{subfigure}
      \caption{original dataset constructed by eq.~(\ref{eq:sechPro}).(Left) shows the $\Re(z(x,t))$ plot for all $x,t$ values at $20000\times 5000$ grid points. (middle) represents the time series plot for two initial conditions. (Right) provides the snapshots of few different time points. Our goal is separate and isolate spatial variables $\Phi_j(x)=j\sech(0.1x+j)$ and $T_j(t)=e^{ijt}$ from the given data constructed by $z(x,t)$.}
        \label{fig:sechExamOrgData}
    \end{figure}
    
\begin{figure}[htb]
\centering
         \begin{subfigure}[b]{0.45\textwidth}
      \centering
        \includegraphics[width=\textwidth]{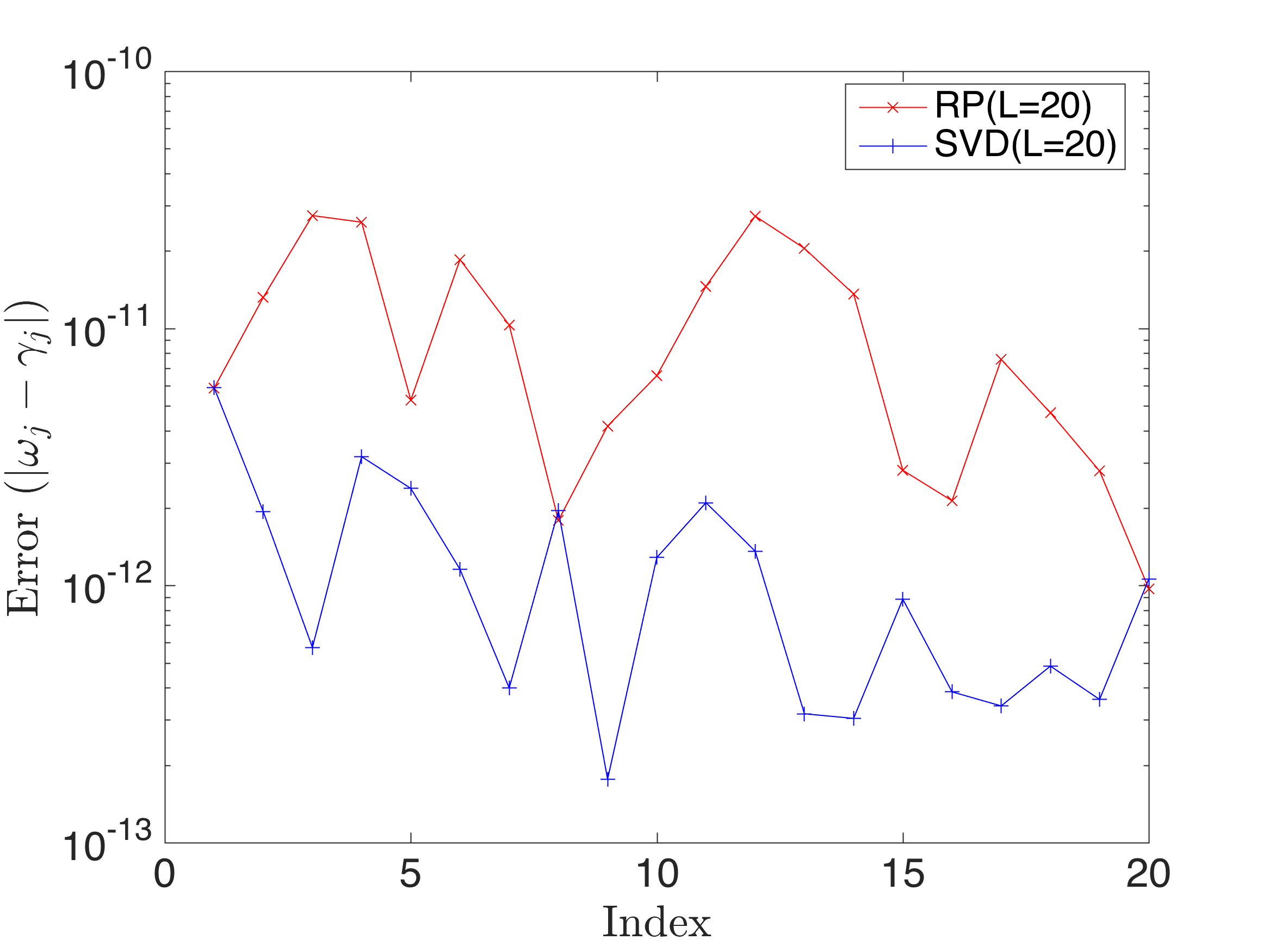}
        \end{subfigure}
                 \begin{subfigure}[b]{0.45\textwidth}
      \centering
        \includegraphics[width=\textwidth]{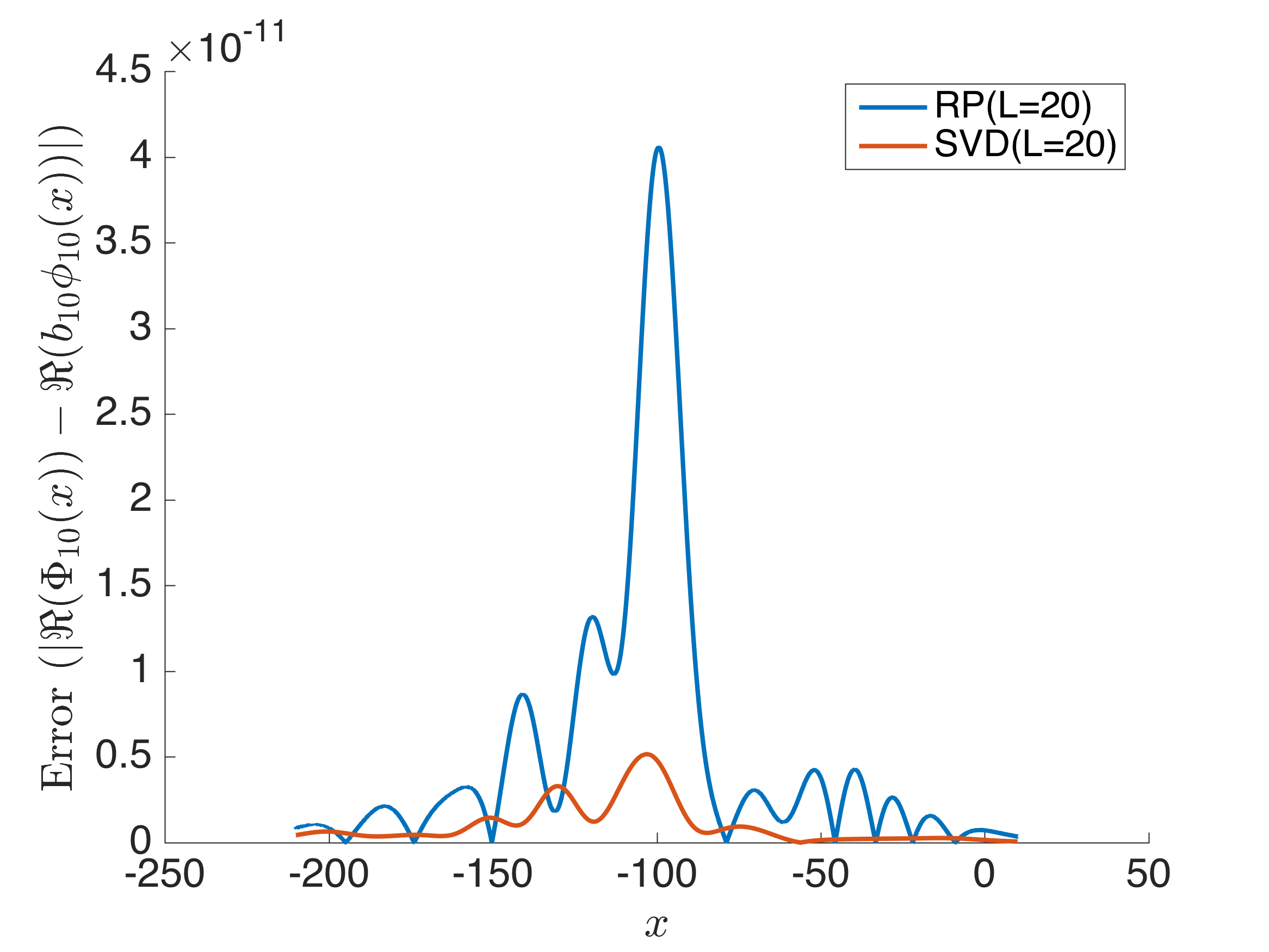}
        \end{subfigure}
        \caption{ (Left) shows the absolute error for estimated eigenvalues from rDMD and exact DMD when the dimension of projected space L=20. (Right) shows the absolute error for estimated 10\textsuperscript{th} DMD mode by rDMD and exact DMD methods. In this case, both methods have very accurate results and error is less than $10^{-10}$. }
        \label{fig:SechSecp}
    \end{figure}

   Comparing this Eq.~(\ref{eq:sechPro}) with decomposition Eq.~(\ref{eq:OrgDec}), the rDMD algorithm is expected to isolate 20 periodic modes by rDMD algorithm. The data set (snapshot matrix) for this problem is constructed by $N=20000$ spatial grid points and  $M=5001$ temporal grid points with $\gamma_j=j$ (see fig.~(\ref{fig:sechExamOrgData})). As discussed in the previous section, if $L\ge 20$ then those expected modes can be isolated and there exist eigenvalues $\lambda_j$ of rDMD operator such that 
    \begin{align*}
    \omega_j=\ln(\lambda_j)=ji
    \end{align*}
    for $j=1,\ 2,\ \dots ,\ 20$ (see fig.~(\ref{fig:SechSecp})). Furthermore, we are expecting corresponding rDMD modes equal to spatial variables of the model such that 
    \begin{align*}
    b_j(x_0)\phi_j(x)=j \sech(0.1x+j)=\Phi_j(x).
    \end{align*}
    As expected, we noticed that calculated modes have negligible error when the dimension of projected space $L\ge r=20$.  Figure \ref{fig:SechSecp} shows the absolute error of eigenvalues and DMD modes. (All the modes behave similarly and here we present mode 10 for demonstration purpose.) by SVD based exact DMD method and random projection based rDMD method. Notice that errors of both methods are less than $10^{-10}$ when $L\ge r=20$. 
    
\begin{figure}[htb]
\centering
         \begin{subfigure}[b]{0.45\textwidth}
      \centering
        \includegraphics[width=\textwidth]{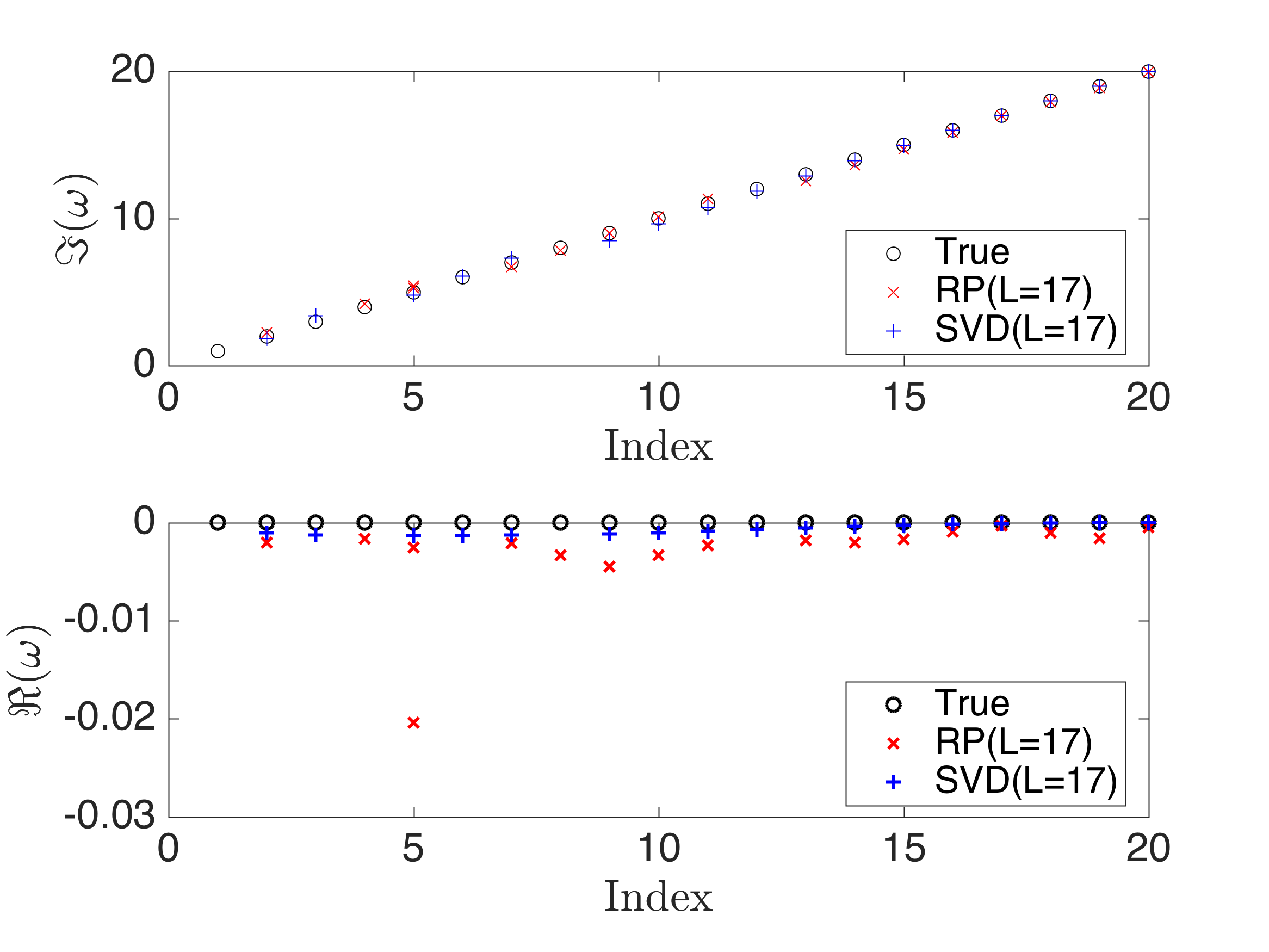}
        \end{subfigure}
                 \begin{subfigure}[b]{0.45\textwidth}
      \centering
        \includegraphics[width=\textwidth]{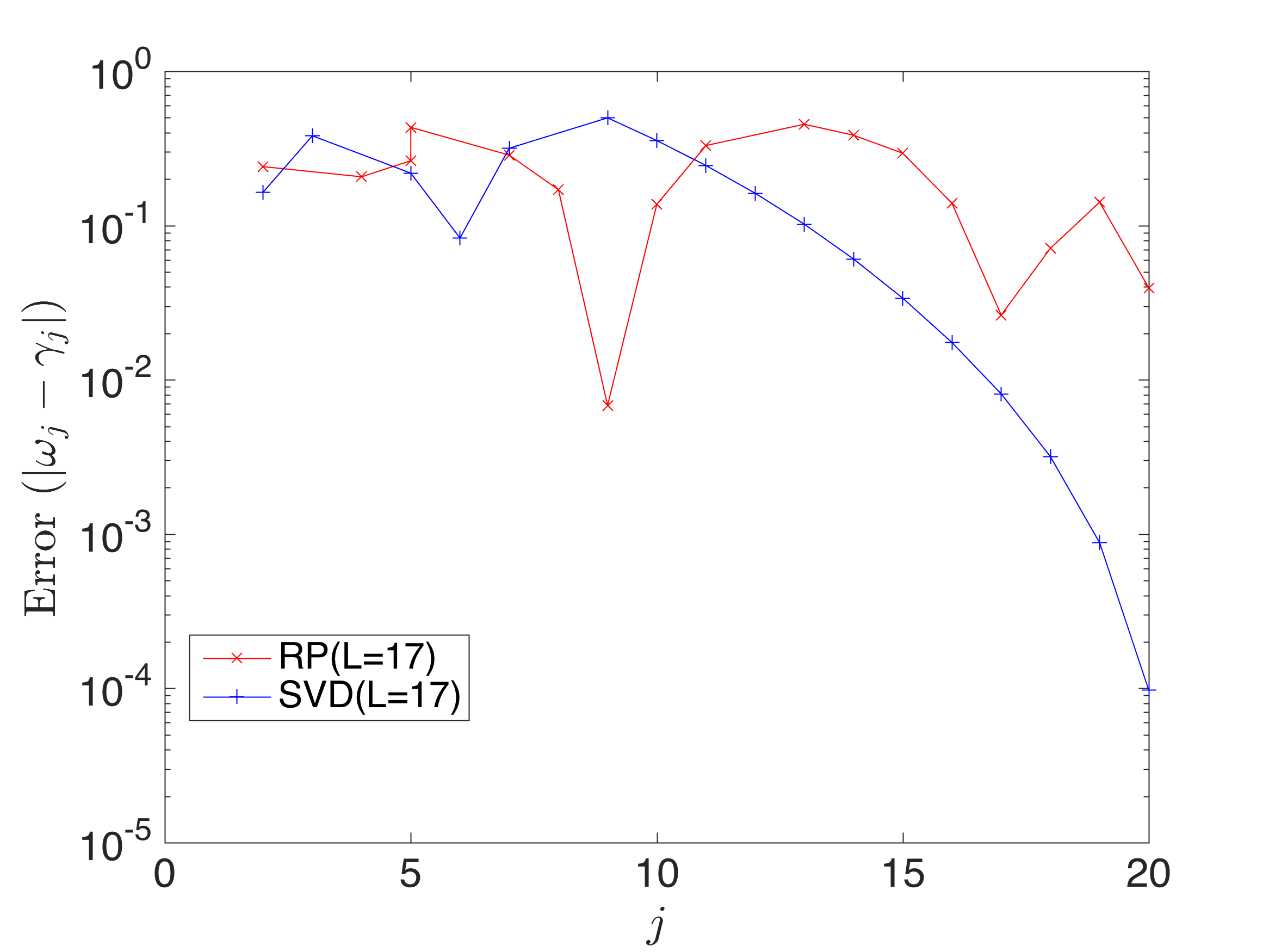}
        \end{subfigure}
        \caption{ (Left) compares the eigenvalues $\omega_j=\ln\lambda_j$ calculated from rDMD (random projection(RP) with L=17) and exact DMD (SVD projection with L=17) methods with the expected true values $\gamma_j=ji$. Here $L=17< 20$ is the dimension of projected space. (Right) shows the absolute error for estimated eigenvalues from rDMD and exact DMD.}
        \label{fig:SechSecpL17}
    \end{figure}

\begin{figure}[htb]
\centering
         \begin{subfigure}[b]{0.45\textwidth}
      \centering
        \includegraphics[width=\textwidth]{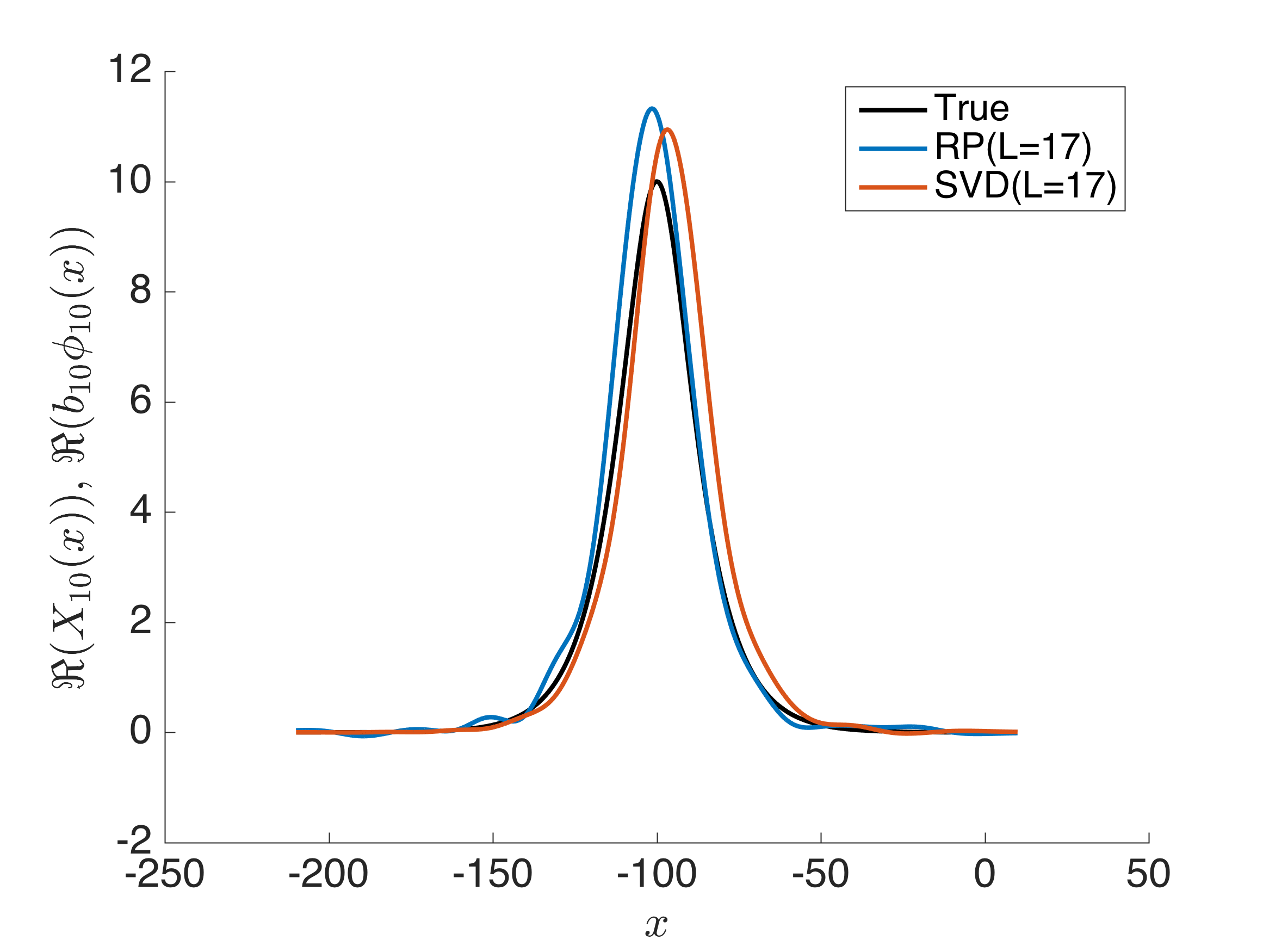}
        \end{subfigure}
                 \begin{subfigure}[b]{0.45\textwidth}
      \centering
        \includegraphics[width=\textwidth]{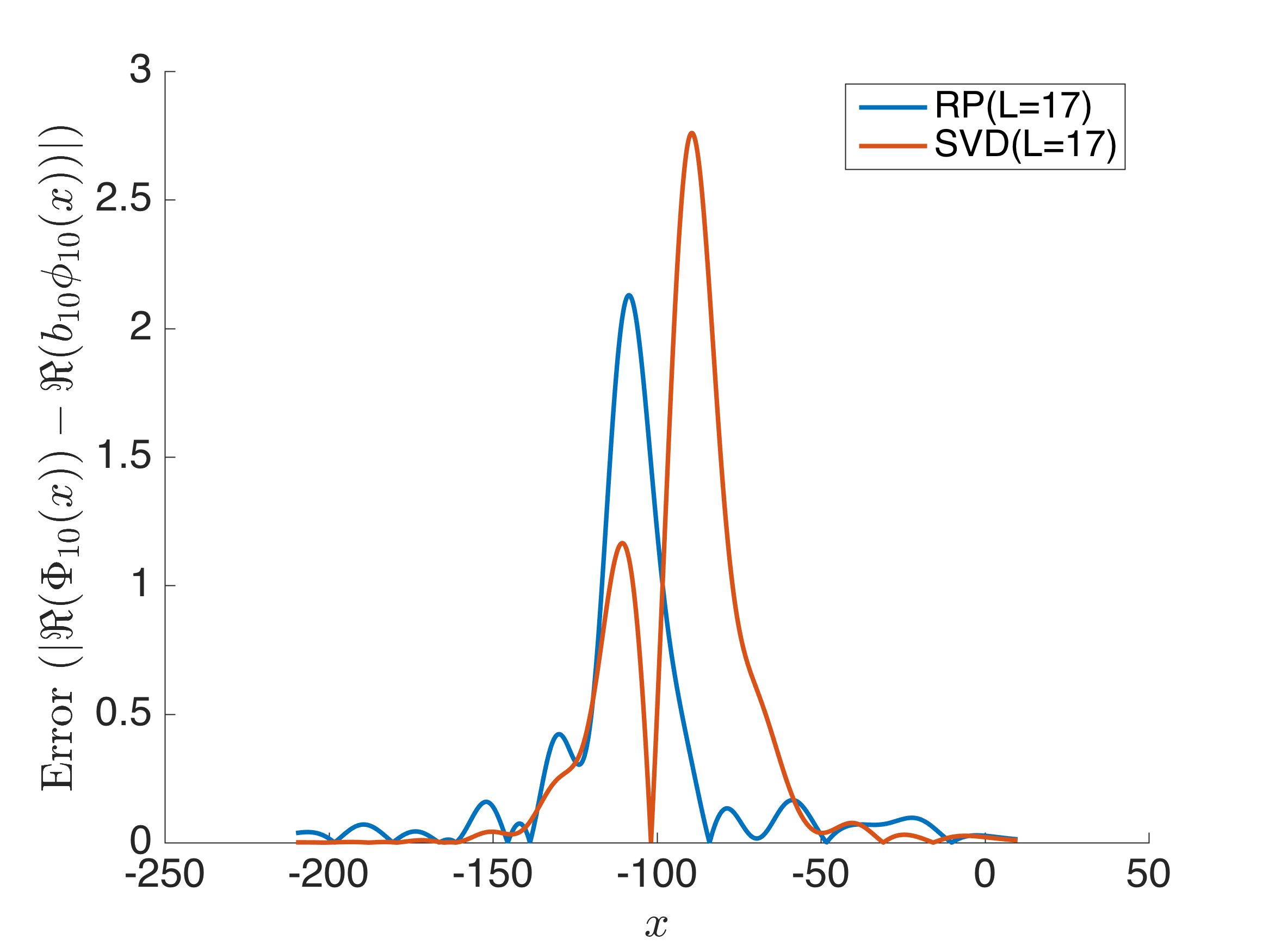}
        \end{subfigure}
        \caption{ (Left) compares  the modes $b_{10}\phi_{10}(x)$ calculated from rDMD (random projection(RP) with L=17) and exact DMD (SVD projection with L=17) methods with the expected true values $\Phi_{10}(x)=10\sech(0.1x+10)$.  (Right) shows the absolute error for estimated values from rDMD  and exact DMD. }
        \label{fig:Mode1_Error_signal}
    \end{figure}

    \begin{table}[ht!]
\centering
\rowcolors{1}{}{white!90!black}
\begin{tabular}{ p{2cm} c c  }
\hline
 {Method }& {Projected by} &{Computational Time(\si{\second}) } \\
 \hline
Exact DMD & SVD & 521.09\\
rDMD & Random Projection & 2.35 \\
 \hline
\end{tabular}
\caption{Computational cost for the SVD based exact DMD and random projection based rDMD method for the data simulated by Eq.~(\ref{eq:sechPro}). Computational cost of SVD for high dimensional snapshot matrix is relatively larger than random projection.}\label{table:Comp_Time_Sig}
\end{table}
      Further, we examine the case when the projected dimension $L=17<r=20$ and compared the results of rDMD with the exact DMD.  We can notice that both methods demonstrate similar errors and rDMD is almost good as the SVD projection based Exact DMD(See Fig.~(\ref{fig:SechSecpL17}) and (\ref{fig:Mode1_Error_signal})). When the number of actual modes($r$) is larger than the dimension of the projected space ($L$), the projected DMD operator only estimates the $L$ number of modes, leads to both truncation errors and error for eigenpair estimation based on projected DMD operator.  The $L<r$ case can be modeled as,
      \begin{align}
          z(x,t)=\sum_{j=1}^{L}\hat{b}_j\hat{\phi}_j(x)e^{\hat{\omega}_jit}+E_{L+1}
      \end{align}
      Where $E_{L+1}=\sum_{L=j+1}^{m}b_j\phi_j(x)e^{\omega_j it}$ is the truncated error that also affects the estimation process of eigenpairs. Therefore, if $L<r$ then there exists an error in eigenvalues and eigenvectors calculated by any method based on the projected DMD. However, this example demonstrates that rDMD can provide the results as good as the SVD projection based method with very low computational cost(See Table.~(\ref{table:Comp_Time_Sig})).  
      
    \subsection{Gulf of Mexico}
    In this example we will consider the data from HYbrid Coordinate Ocean Model (HYCOM) \cite{hycom} which simulates the ocean data around the Gulf of Mexico. We used hourly surface velocity component ($u,v$) with $1/25^0$  spatial resolution ($N=541\times 347$ grid points) data for 10 days ($240$ \si{\hour} and $M=239$). Understanding the dynamics from the oceanographic data is an interesting application of DMD because those dynamics can be decomposed by tidal constituents %associated with the tidal frequencies and data used for this analysis is 
   . Hence, we are expected to isolate the dynamics associated with the tidal period ; in other words, the final DMD mode selection is based on the period $P_i=2\pi/\Im(\ln(\lambda_i))$ of the modes(see table (\ref{table:GOMmodes})). We constructed the snapshot matrix 
    \begin{align}
        X=\begin{bmatrix} u\\
        v
        \end{bmatrix}
    \end{align}
   by stacking the snapshots of velocity components ($u,v$)  in each column to perform the DMD analysis.
    
    \begin{table}[ht!]
\centering
\rowcolors{1}{}{white!90!black}
\begin{tabular}{ p{1cm} c c c }
\hline
 {Mode}& \multicolumn{2}{c}{Period(\si{\hour})} &  {Associated Feature}  \\
 {} & {DMD} &{rDMD}&{}\\
 \hline
1& $\infty$ & $\infty$ &Gulf stream around the GOM.(see fig.~(\ref{fig:GOMModes1})) \\
2& $12.47$& $12.47$ & Semi-diurnal tidal constituents.(see fig.~(\ref{fig:GOMModes2})) \\
3& $23.85$& $24.56$ & Diurnal tidal constituents.(see fig.~(\ref{fig:GOMmode3to5})) \\
4& $6.08$& $6.07$& 2\textsuperscript{nd} harmonic to semi-diurnal tidal constituents. (see fig.~(\ref{fig:GOMmode3to5})) \\
5& $4.16$ & $4.17$ &3\textsuperscript{rd} harmonic to semi-diurnal tidal constituents. (see fig.~(\ref{fig:GOMmode3to5})) \\
 \hline
\end{tabular}
\caption{DMD modes for Gulf of Mexico data set. Modes are selected based on the association to the tidal periods. }\label{table:GOMmodes}
\end{table}

        \begin{figure}[htb]
        \centering
         \begin{subfigure}[b]{0.45\textwidth}
      \centering
        \includegraphics[width=\textwidth]{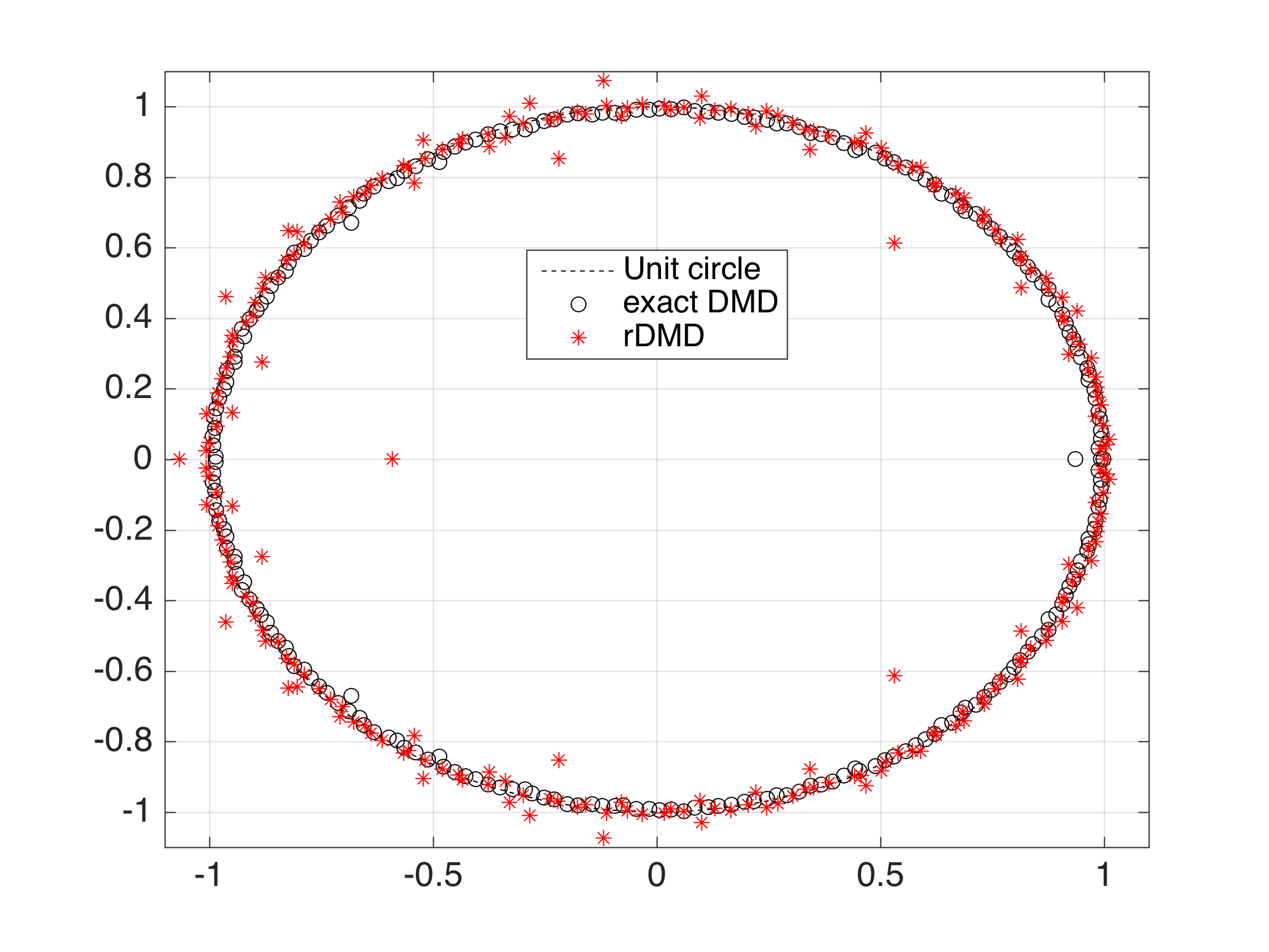}\caption{}
        \end{subfigure}
                 \begin{subfigure}[b]{0.45\textwidth}
      \centering
        \includegraphics[width=\textwidth]{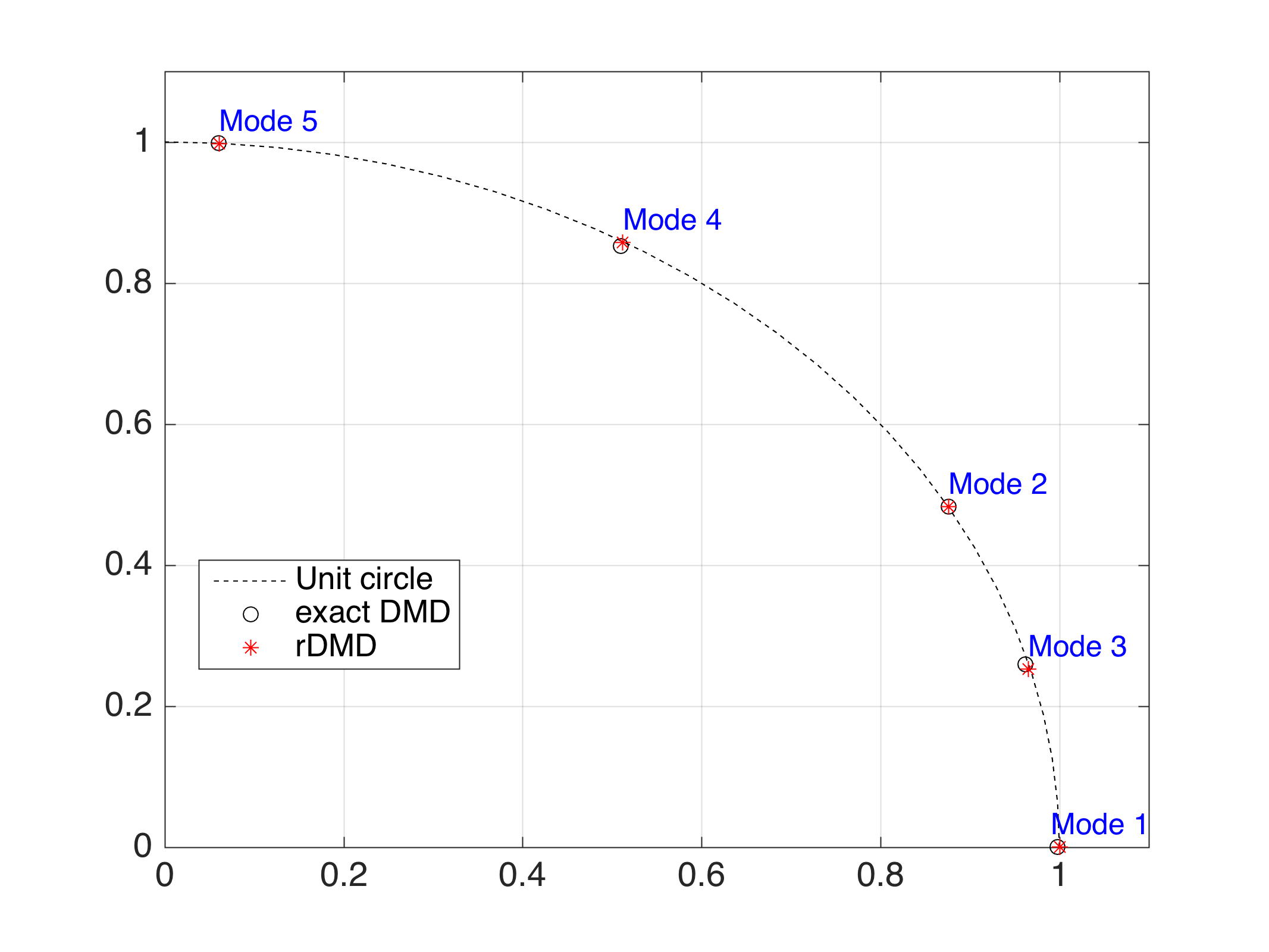}\caption{}
        \end{subfigure}
        \caption{ Eigenvalues $\lambda_i$ calculated from exact DMD and rDMD methods. (a) Full spectrum of the two methods with projected space dimension $L=239$ and (b) shows the first 5 modes. The mode selection is based on the comparison of the tidal periods with period of the DMD modes.}
        \label{fig:GOM_Eigenvalue}
    \end{figure}
    
\begin{figure}[ht!]
        \centering
         \begin{subfigure}[b]{0.45\textwidth}
      \centering
        \includegraphics[width=\textwidth]{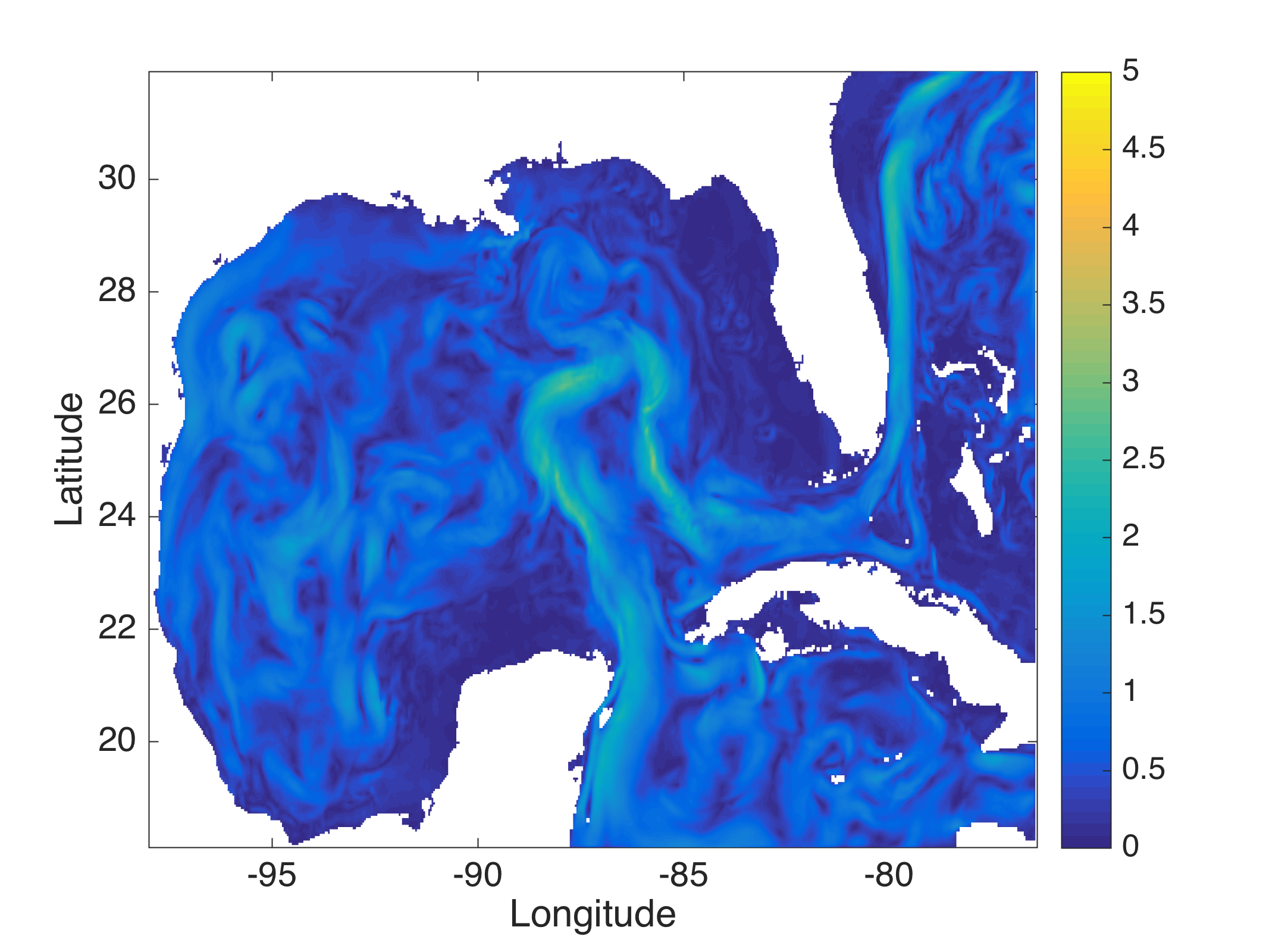}\caption{}
        \end{subfigure}
                 \begin{subfigure}[b]{0.45\textwidth}
      \centering
        \includegraphics[width=\textwidth]{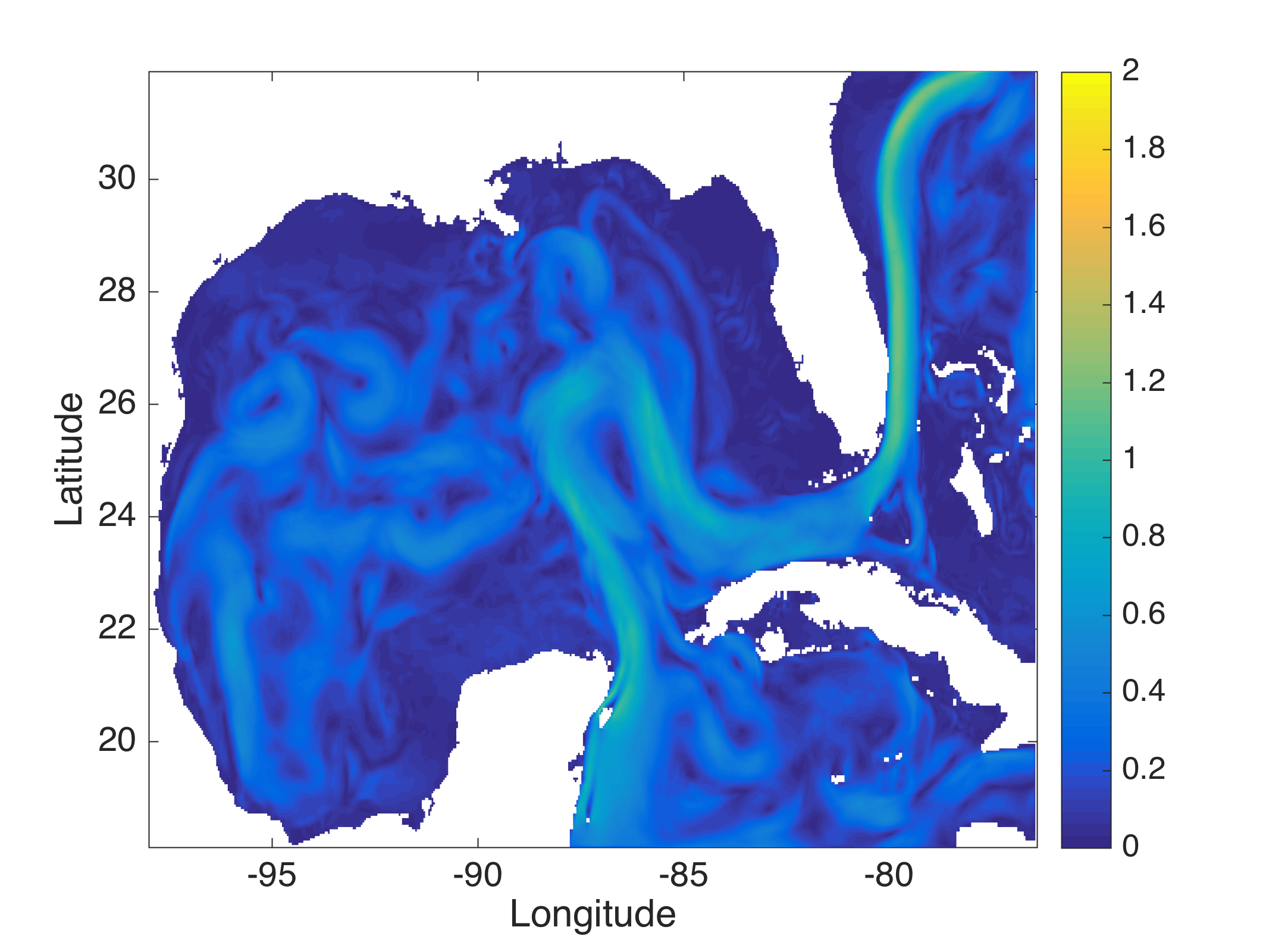}\caption{}
        \end{subfigure}
        \caption{This figure compares the (a) DMD  and (b) rDMD background mode identified by data from the Gulf of Mexico (GOM). This background mode captures the ocean current passing through the GOM. }
        \label{fig:GOMModes1}
    \end{figure}

\begin{figure}[ht!]
        \centering
         \begin{subfigure}[b]{0.45\textwidth}
      \centering
        \includegraphics[width=\textwidth]{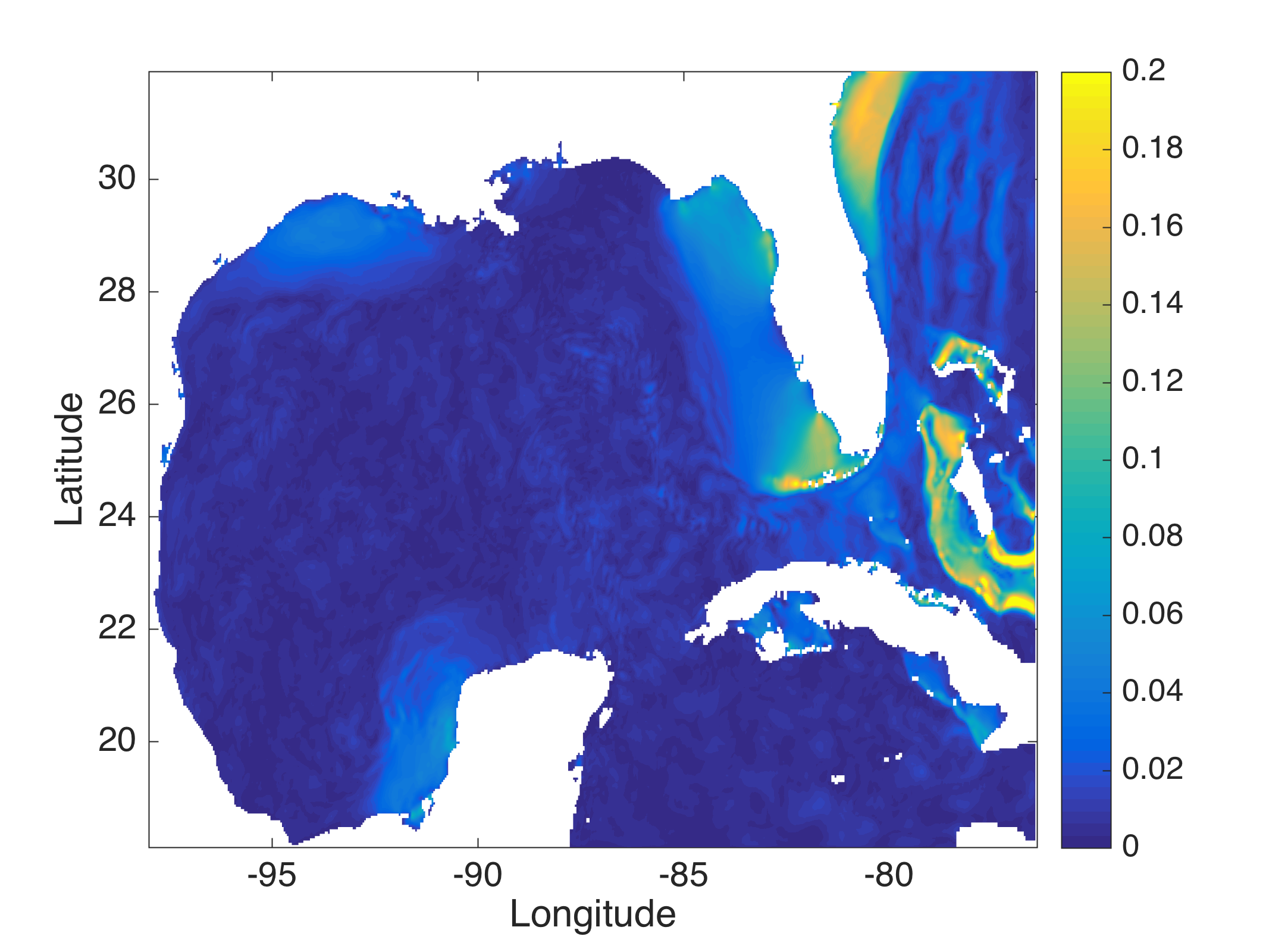}\caption{}
        \end{subfigure}
                 \begin{subfigure}[b]{0.45\textwidth}
      \centering
        \includegraphics[width=\textwidth]{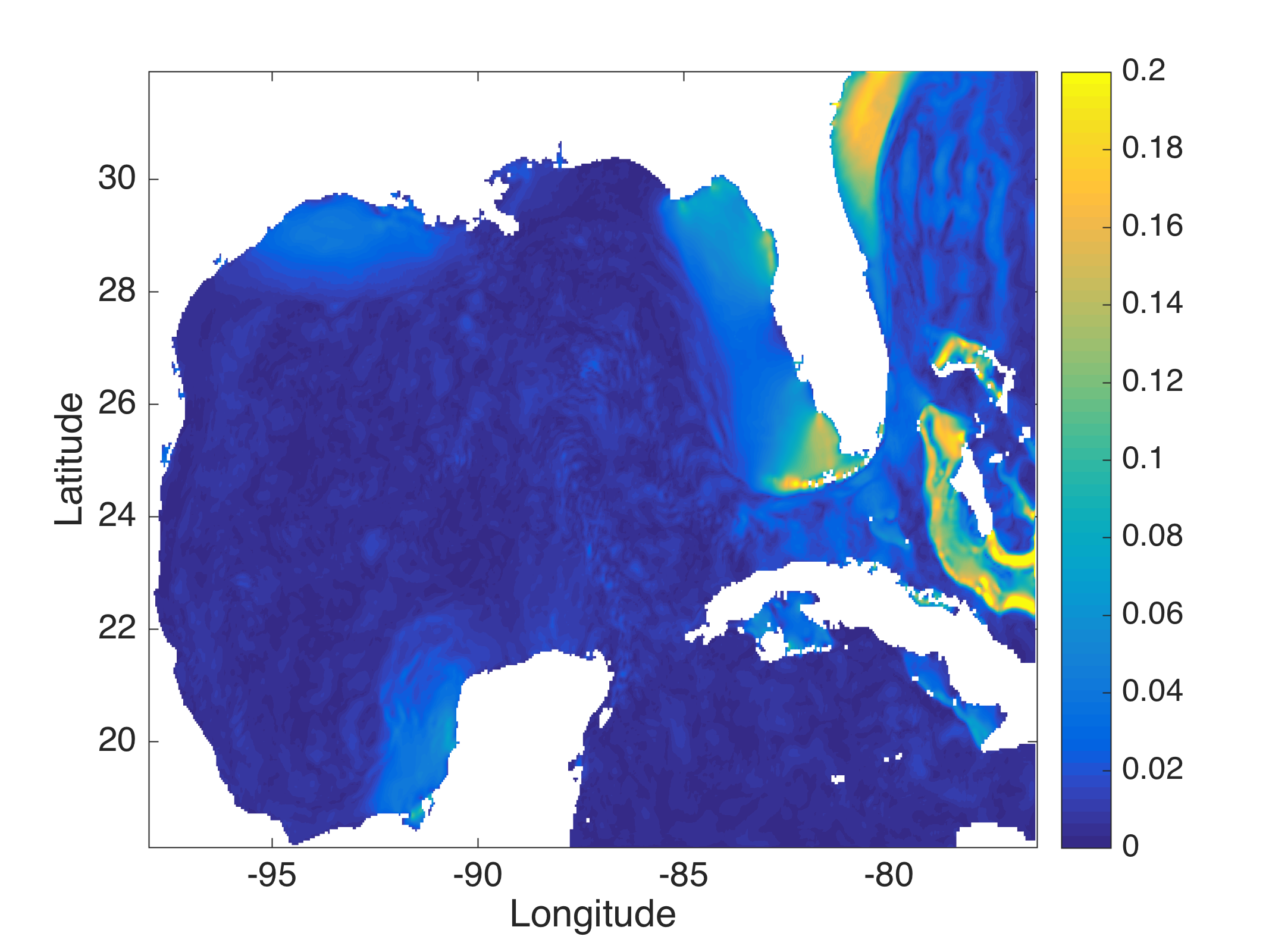}\caption{}
        \end{subfigure}
        \caption{This figure compere the (a) DMD  and (b) rDMD mode associated with M2 tidal frequency. This mode capture the "red tides".  }
        \label{fig:GOMModes2}
    \end{figure}

\begin{figure}[ht!]\centering
  \begin{subfigure}{0.32\textwidth}
    \includegraphics[width=\linewidth]{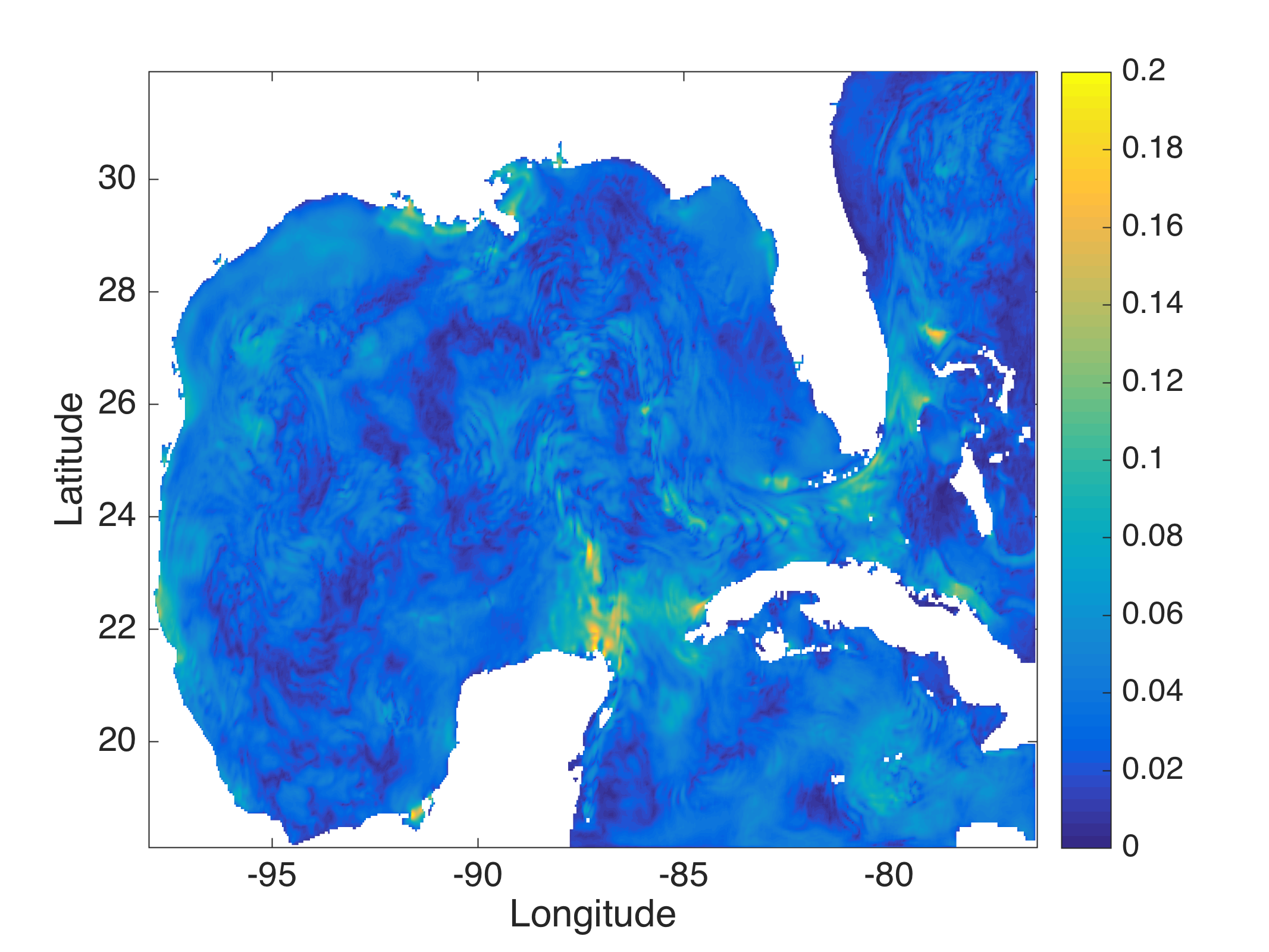}
    \label{fig:GOMmod3svd}
  \end{subfigure}
\begin{subfigure}{0.32\textwidth}
    \includegraphics[width=\linewidth]{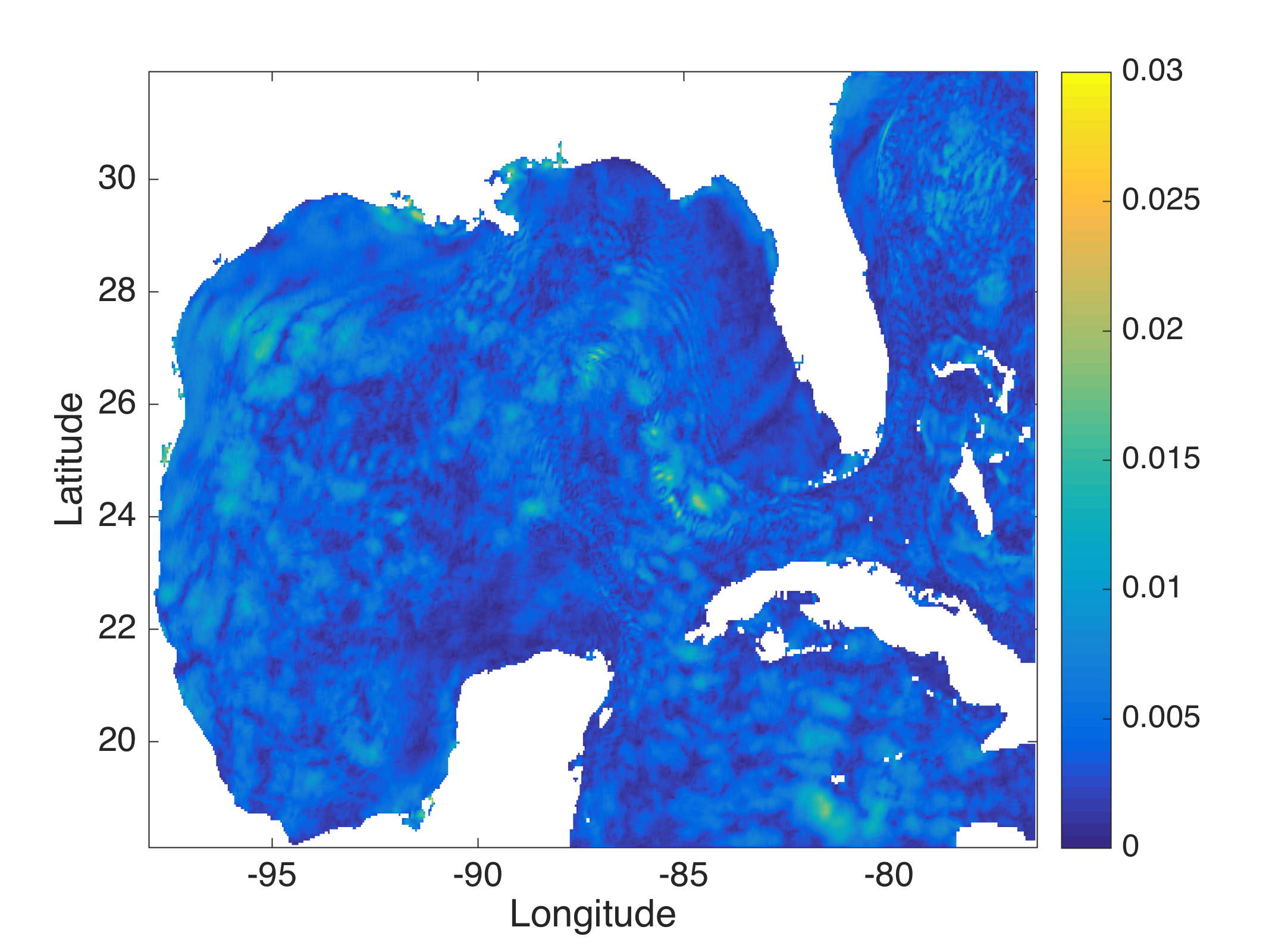}
    \label{fig:GOMmod4svd}
  \end{subfigure}
  \begin{subfigure}{.32\textwidth}
    \includegraphics[width=\linewidth]{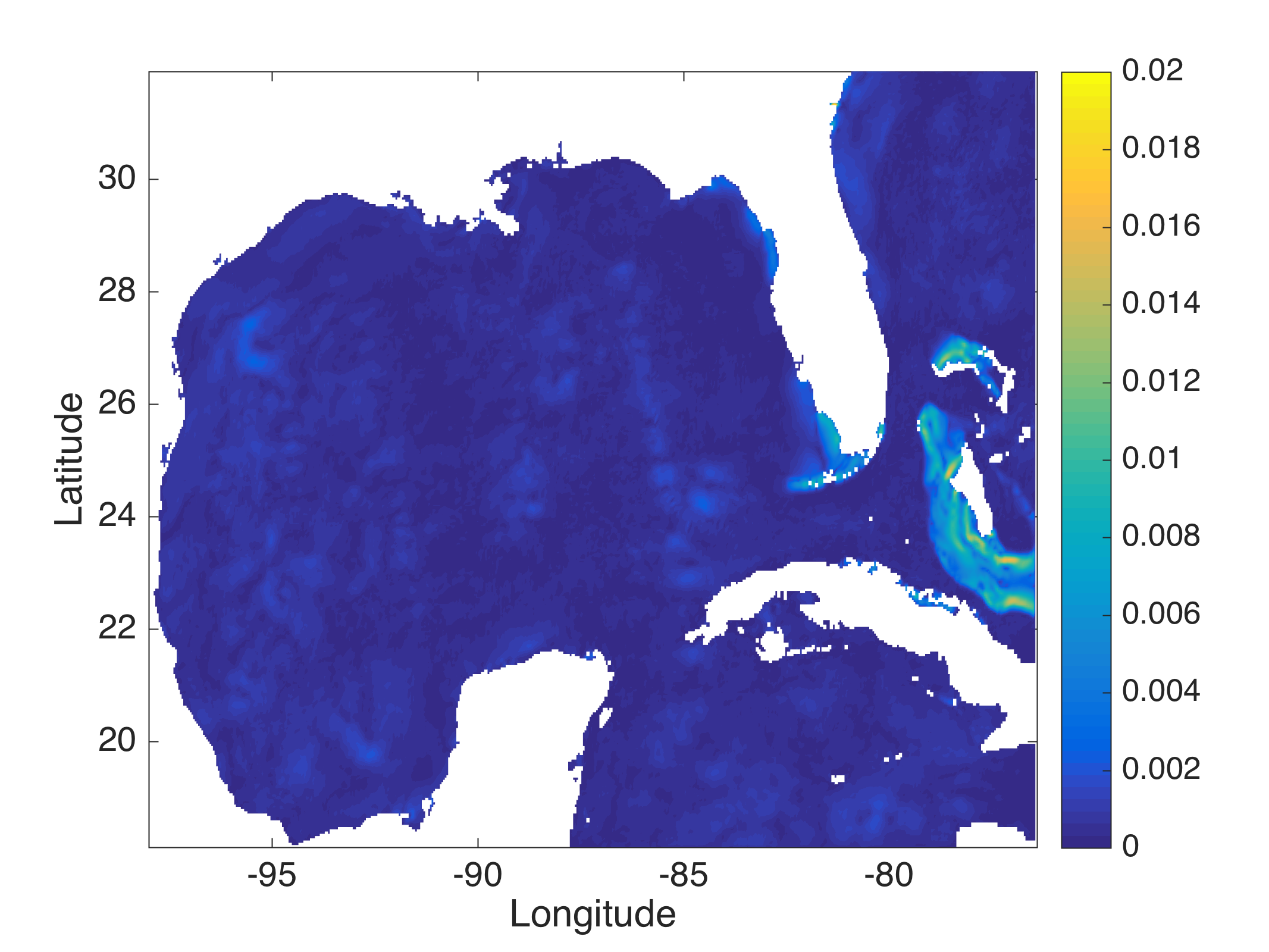}
   \label{fig:GOMmod5svd}
  \end{subfigure}

   \begin{subfigure}{0.32\textwidth}
    \includegraphics[width=\linewidth]{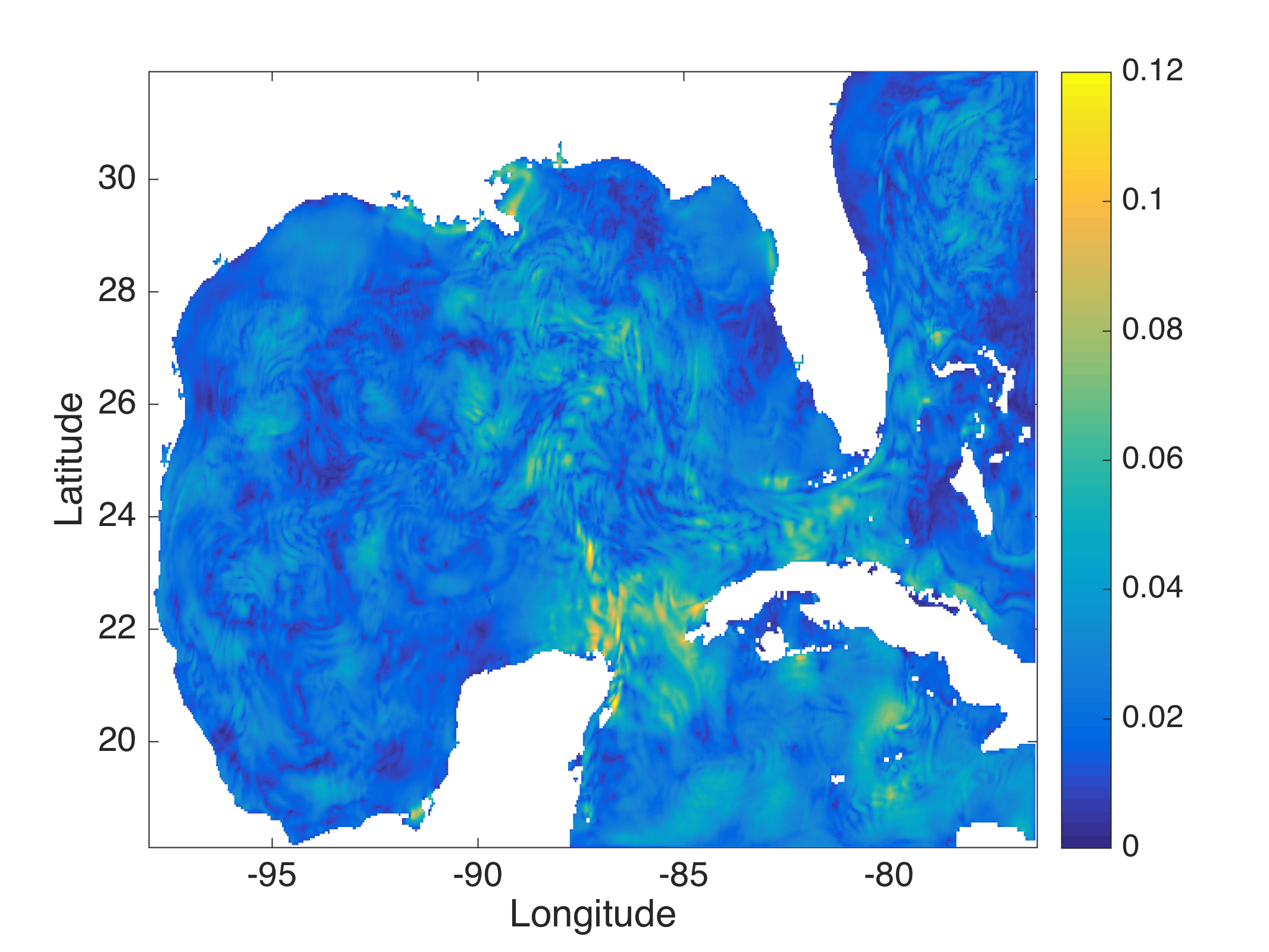}
    \label{fig:GOMmod3r}
  \end{subfigure}
\begin{subfigure}{0.32\textwidth}
    \includegraphics[width=\linewidth]{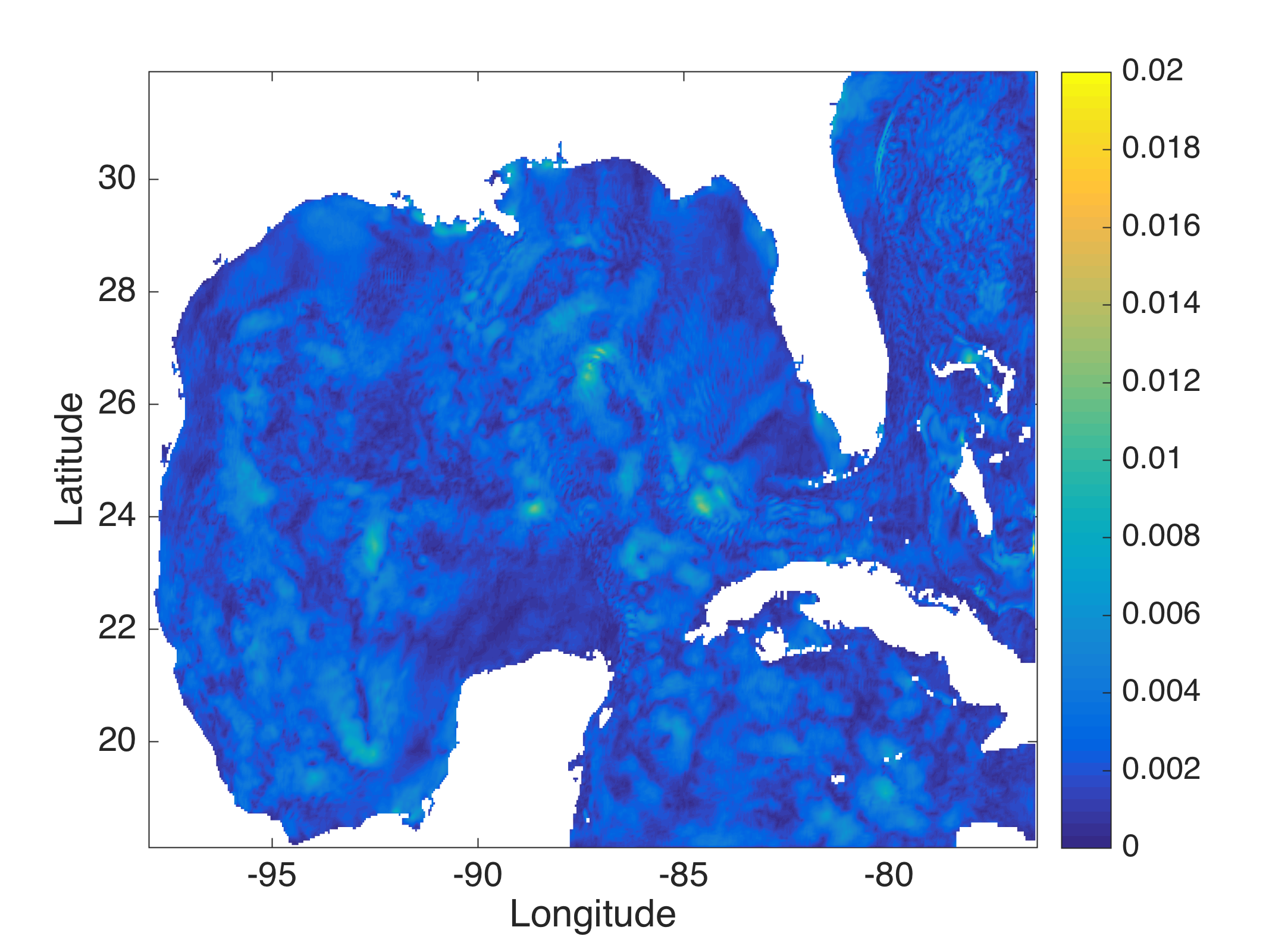}
    \label{fig:GOMmod4r}
  \end{subfigure}
  \begin{subfigure}{.32\textwidth}
    \includegraphics[width=\linewidth]{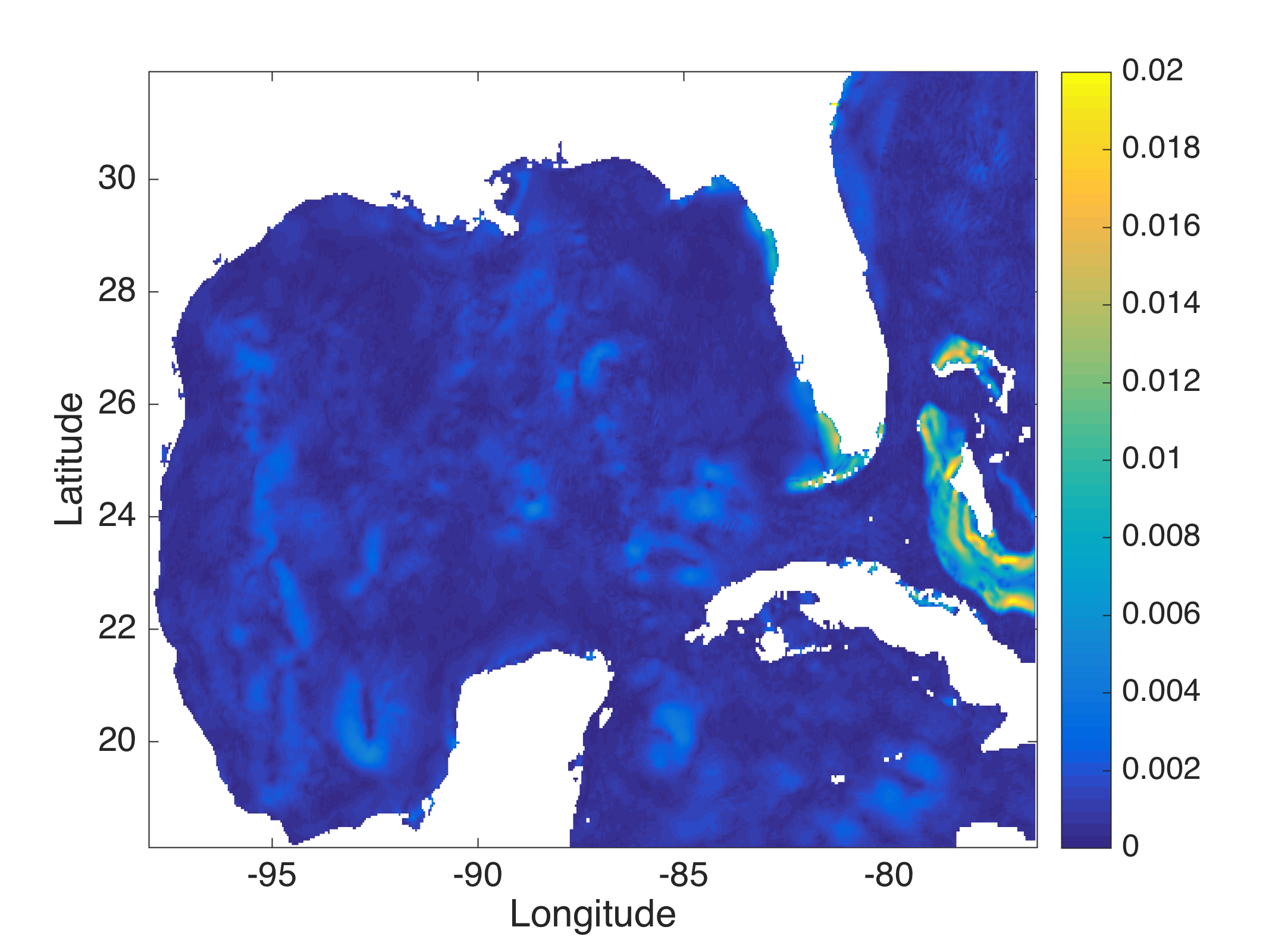}
    \label{fig:GOMmod5r}
  \end{subfigure}
  \caption{First row represent the exact DMD modes 3, 4 and 5 (left to right) and second row shows the rDMD modes 3, 4 and 5 ((left to right)). Mode 3 is a diurnal mode with period $23.85$ \si{hour} for exact DMD case and $24.56$\si{hour} for rDMD case. Mode 4 and 5 are associated with 2\textsuperscript{nd} and 3\textsuperscript{rd} harmonic of a semi-diurnal tidal constituents respectively.}\label{fig:GOMmode3to5}
\end{figure}

    Figure (\ref{fig:GOM_Eigenvalue}) shows that most of the eigenvalues calculated from SVD based exact DMD and random projection based rDMD are agree. Furthermore, eigenvalues that isolated the specific dynamics are almost equal. Additionally, figure (\ref{fig:GOMModes1})-(\ref{fig:GOMmode3to5}) shows the spacial profile of those modes from exact DMD and rDMD methods. Also, each mode clearly isolated the interesting oceanographic features(see table (\ref{table:GOMmodes})) and both methods provide almost the same spacial structures(see fig.~(\ref{fig:GOMModes1})-(\ref{fig:GOMmode3to5})) as expected. 
    
    Notice that the dimension of the snapshot matrix is $375454 \times 239$ and SVD calculation of this matrix is more costly for both computation and storage.  On the other hand, random projection performs only by matrix multiplication which can be done at relatively low cost. Hence, we can achieve almost the same results by using random projection method at a relatively much lower computational and storage cost.

\section{Conclusion}
We have demonstrated that our rDMD can achieve very accurate results with low-dimensional data embedded in high-dimensional observable space. Recent analytic technology from the concepts of high-dimensional geometry of data, and concentration of measure have born out that perhaps surprising if not initially intuitively that even random projection methods can be quite powerful and capable. Here, in the setting of DMD methods approximating and projecting the action of a Koopman operator, we show that randomized projection can be developed and analyzed rigorously by the Johnson-Lindenstrass theorem formalism, this showing a powerful and simple approach.
We provided a theoretical framework and experimental results to address those issues raised from SVD based methods by introducing our new rDMD algorithm. The theoretical framework is based on generalizing the SVD based concept as a projection of high dimensional data into a low dimensional space. We proved that eigenpairs of DMD in original space can be estimated by using any rank $L$ projection matrix $P$. Being able to estimate eigenpairs allowed us to use the powerful and simple  Johnson-Lindenstrauss lemma and the random projection method allowing us to project data with matrix multiplication. Therefore our proposed random projection-based DMD(rDMD) can estimate eigenpairs of the DMD operator with low storage and computational cost. Further, the error of the estimation can be controlled by choosing the dimension of the projected space and we demonstrated this error bound through the "logistic map" example. 

DMD promises the separation of the spatial and time variables from data. Hence, we experimentally demonstrated how well the rDMD algorithm performed this task by a toy example. Notice that the number of those isolated modes ($m$) are relatively (to spatial and temporal resolution) low in practical applications.  If $m<< M$, then the rank of the data matrix is much lower, and those eigenvalues and vectors of interest can be  estimated accurately by projecting data into the much lower dimensional space $L\ge m$. The SVD projection-based exact DMD method still needs to calculate the SVD of a high dimensional(roughly $10^{10}\times 10^{3}$ ) data matrix while rDMD only requires to multiply the data matrix by a much lower-dimensional projection matrix. Furthermore, we noticed that both exact and random DMD methods are experiencing similar errors. However random projection is much faster and needs less space for the calculations. We also demonstrate that practical applications also provide similar results by using oceanographic data from the Gulf of Mexico.

Since the size of the DMD matrix is enormous in those applications  (this could be roughly $10^{10}\times 10^{10}$), the eigenpairs of it must be estimated by projecting data into low dimensional space. Estimating eigenvalues and eigenvectors of a DMD operator using a high dimensional snapshot data matrix (in applications this could be $10^{10}\times 10^{3}$) with existing SVD based methods is expensive. The computational efficiency of the rDMD led to a new path of current Koopman analysis. It allows using more observable variables in the data matrix without need of much extra computational power. Hence, state variables and more non-linear terms of them can be used in analysis with low cost to improve the Koopman modes. JL theory can be adopted further into the field of numerical methods of Koopman theory. As a next step, we can use the random projection concept in the extended DMD and kernel DMD methods.

\vspace{6pt} 

%%%%%%%%%%%%%%%%%%%%%%%%%%%%%%%%%%%%%%%%%%
%% optional
%\supplementary{The following are available online at \linksupplementary{s1}, Figure S1: title, Table S1: title, Video S1: title.}


%%%%%%%%%%%%%%%%%%%%%%%%%%%%%%%%%%%%%%%%%%
%\authorcontributions{Conceptualization, Sudam Surasinghe and Erik M. Bollt; Data curation, Sudam Surasinghe and Erik M. Bollt; Formal analysis, Sudam Surasinghe and Erik M. Bollt; Funding acquisition, Erik M. Bollt; Methodology, Sudam Surasinghe and Erik M. Bollt; Project administration, Erik M. Bollt; Resources, Erik M. Bollt; Software, Sudam Surasinghe and Erik M. Bollt; Supervision, Erik M. Bollt; Validation, Sudam Surasinghe and Erik M. Bollt; Visualization, Sudam Surasinghe and Erik M. Bollt; Writing – original draft, Sudam Surasinghe and Erik M. Bollt; Writing – review \& editing, Sudam Surasinghe and Erik M. Bollt.}

%%%%%%%%%%%%%%%%%%%%%%%%%%%%%%%%%%%%%%%%%%
\funding{EB gratefully acknowledges funding from the Army Research Office W911NF16-1-0081 (Dr Samuel Stanton) as well as from DARPA.}

%%%%%%%%%%%%%%%%%%%%%%%%%%%%%%%%%%%%%%%%%%
%\acknowledgments{ }

%%%%%%%%%%%%%%%%%%%%%%%%%%%%%%%%%%%%%%%%%%
%\conflictsofinterest{Declare conflicts of interest or state ``The authors declare no conflict of interest.'' Authors must identify and declare any personal circumstances or interest that may be perceived as inappropriately influencing the representation or interpretation of reported research results. Any role of the funders in the design of the study; in the collection, analyses or interpretation of data; in the writing of the manuscript, or in the decision to publish the results must be declared in this section. If there is no role, please state ``The funders had no role in the design of the study; in the collection, analyses, or interpretation of data; in the writing of the manuscript, or in the decision to publish the results''.} 

%%%%%%%%%%%%%%%%%%%%%%%%%%%%%%%%%%%%%%%%%%
%% optional
%\abbreviations{The following abbreviations are used in this manuscript:\\

%\noindent 
%\begin{tabular}{@{}ll}
%MDPI & Multidisciplinary Digital Publishing Institute\\
%DOAJ & Directory of open access journals\\
%TLA & Three letter acronym\\
%LD & linear dichroism
%\end{tabular}}

%%%%%%%%%%%%%%%%%%%%%%%%%%%%%%%%%%%%%%%%%%
%% optional
\appendixtitles{yes} %Leave argument "no" if all appendix headings stay EMPTY (then no dot is printed after "Appendix A"). If the appendix sections contain a heading then change the argument to "yes".
%\appendix
%\section{}
\unskip

%\section{}
%All appendix sections must be cited in the main text. In the appendixes, Figures, Tables, etc. should be labeled starting with `A', e.g., Figure A1, Figure A2, etc. 


%=====================================
% References, variant A: internal bibliography
%=====================================
\reftitle{References}
% The following MDPI journals use author-date citation: Arts, Econometrics, Economies, Genealogy, Humanities, IJFS, JRFM, Laws, Religions, Risks, Social Sciences. For those journals, please follow the formatting guidelines on http://www.mdpi.com/authors/references
% To cite two works by the same author: \citeauthor{ref-journal-1a} (\citeyear{ref-journal-1a}, \citeyear{ref-journal-1b}). This produces: Whittaker (1967, 1975)
% To cite two works by the same author with specific pages: \citeauthor{ref-journal-3a} (\citeyear{ref-journal-3a}, p. 328; \citeyear{ref-journal-3b}, p.475). This produces: Wong (1999, p. 328; 2000, p. 475)

%=====================================
% References, variant B: external bibliography
%=====================================
\externalbibliography{yes}
\bibliography{bibit}

%%%%%%%%%%%%%%%%%%%%%%%%%%%%%%%%%%%%%%%%%%
%% optional
%\sampleavailability{Samples of the compounds ...... are available from the authors.}

%% for journal Sci
%\reviewreports{\\
%Reviewer 1 comments and authors’ response\\
%Reviewer 2 comments and authors’ response\\
%Reviewer 3 comments and authors’ response
%}

%%%%%%%%%%%%%%%%%%%%%%%%%%%%%%%%%%%%%%%%%%
\end{document}